\pdfoutput=1
\documentclass{article}

\usepackage{arxiv}          

\usepackage[utf8]{inputenc} 
\usepackage[T1]{fontenc}    
\usepackage{microtype}      
\usepackage{graphicx}
\usepackage{url}            

\usepackage{amsmath}
\usepackage{amssymb}
\usepackage{amsfonts}       
\usepackage{nicefrac}       

\usepackage{booktabs}       
\usepackage{multirow}
\usepackage{array}
\usepackage{longtable}
\usepackage{threeparttable} 
\usepackage{float}
\usepackage{pdflscape}      
\usepackage{rotating}       

\usepackage{enumitem}
\usepackage{setspace}

\usepackage[labelfont=bf,textfont=bf]{caption}
\usepackage{subcaption}

\usepackage{xcolor}
\usepackage{tikz}
\usetikzlibrary{arrows.meta, positioning, shadows, patterns, shapes.geometric}

\usepackage[numbers]{natbib}
\usepackage{doi}

\usepackage{hyperref}       

\title{\textbf{Tabular Deep Learning for Algorithmic Trading: Cross-Regime Bayesian 
                Optimisation for Equity Signal Generation}}

\author{\hspace{1mm}Josh Le Grice\thanks{This work was conducted as a part of the MSc in Data Science at the University of Exeter}\\
	Department of Computer Science\\
	University of Exeter\\
	\texttt{joshualegrice@gmail.com} \\
}

\hypersetup{
	pdftitle={Tabular Deep Learning for Algorithmic Trading: Cross-Regime Bayesian 
                Optimisation for Equity Signal Generation},
	pdfsubject={cs.LG, cs.AI, cs.CY},
	pdfauthor={Josh Le Grice},
	pdfkeywords={algorithmic trading, deep learning, alternative data, financial markets},
}

\begin{document}
\maketitle

\begin{abstract}
	Algorithmic trading now represents a market exceeding \$20 billion, 
where even marginal gains in signal robustness can translate into 
economically significant returns. Existing evaluations of equity 
prediction models do not explicitly target regime robustness during 
hyperparameter selection. Five 
model classes are trained on daily observations from approximately 
300 large-cap US equities over eleven years, with 
Bayesian optimisation configured to target trading performance 
across three statistically different market regimes. 
Regime-robust hyperparameter selection is associated with 
out-of-sample generalisation, as signal precision remains above 
the random baseline across all four quarters of the test period, 
and portfolio performance slowly degrades under simulated input 
noise before collapsing beyond a defined threshold. No individual 
tabular deep learning architecture outperforms 
gradient-boosted trees, but combining XGBoost and TabNet using 
rank aggregation produces a Hybrid ensemble with an annualised 
return of $51.26\%$, a Sharpe ratio of $2.44$, and a statistically 
significant CAPM alpha of $0.423$ ($p = 0.011$). 
A near-zero beta indicates this outperformance 
is driven by stock selection, not market exposure. Alternative 
data plays a secondary role once technical and fundamental features 
are accounted for, as well as contributing more strongly on the 
short side than the long, and varies by model class. 
An interactive application makes these results explorable in real 
time, with live data integration the remaining step toward 
practical deployment.

\end{abstract}


\newpage

\section{Introduction}\label{sec:introduction}

Stock markets aggregate information about expected cash flows, risk, and investor 
beliefs, which translate into prices that influence how capital is allocated~\cite{Wurgler1999}. 
Understanding price behaviour dictates how 
efficiently that capital is distributed, altering investment decision-making  
and portfolio construction. More accurate return predictions improve 
these decisions~\cite{Ma2021, Babiak2026}. However, poor signals incur 
costs, including misallocated capital and underperforming portfolios. 

Computational methods and machine learning (ML) have driven the 
development of prediction strategies~\cite{Zhang2025deep, 
GiantsidiSofia2025Dlff, Ma2021}, but regime instability 
frequently limits their profitability. Strategies that 
perform well in one market environment often fail in another 
as conditions shift, making consistent returns difficult 
to sustain~\cite{Timmermann2008, Paye2006}. 
Algorithmic trading automates these strategies into live 
markets~\cite{LiuPeng2023}, and already represents a \$20.23 
billion market in 2026, that is projected to reach \$29.54 billion 
by 2031~\cite{Mordor2026}. Marginal improvements in strategy robustness could therefore 
generate significant alpha. To capture this alpha, a novel cross-regime Bayesian optimisation 
framework is developed, examining whether such a framework produces 
configurations that generalise to unseen market conditions. 
We hypothesise that configurations selected under a cross-regime 
objective retain predictive performance across distinct regimes 
in the OOS period and remain stable under realistic input perturbation.

Cross-sectional equity prediction presents structural 
challenges beyond regime instability. Daily returns are highly stochastic, 
making prediction of continuous return magnitudes unreliable and necessitates a 
cross-sectional ranking approach.
Built from stock-day observations containing firm 
characteristics, market variables, sentiment measures, and macroeconomic 
indicators~\cite{Kumbure2022}, cross-sectional equity prediction is structurally different from time-series 
forecasting~\cite{gu2020}. Deep learning models have 
underperformed tree-based methods on such inputs~\cite{Grinsztajn2022, ShwartzZiv2022}, suggesting architectures 
designed for tabular structure may be more 
appropriate~\cite{Arik2021, Gorishni2021, HuangXin2020, Somepalli2021}, but their performance under 
regime-robust evaluation remains untested in the literature. 
Therefore, in what is, to our knowledge, the first such evaluation, we compare tabular deep learning (TDL) models with traditional ML 
baselines in daily trading signal 
generation and portfolio construction for US markets, the 
hypothesis being that their architectural inductive biases 
translate into consistent gains over tree-based methods.

Generating reliable signals may require source diversification to include alternative data
alongside price and accounting variables. News sentiment and search-based attention 
proxies capture behaviour that conventional variables miss~\cite{SunYunchuan2024, 
Tetlock2007, Preis2013}, and markets respond to attention shocks and sentiment dynamics that fundamental data 
cannot capture. If these sources improve forecasting in ways that hold up under realistic trading 
conditions is poorly understood. This introduces our final question of how alternative data contributes to model
predictions and trading performance, and what SHAP analyses reveal about their relative importance. We 
hypothesise that these sources will account for at least 10\% of mean 
SHAP attribution across both signal predictions, a threshold derived from 
the 18\% variable share documented in the financial ML literature~\cite{Kumbure2022}.
10\% is chosen as a conservative threshold to account for the fact that variable share
is not a direct measure of predictive attribution.

Even where predictive signal exists, trading costs reduce returns 
that look promising under statistical evaluation~\cite{Ma2021,lopezdeprado2018}. 
Predictive accuracy and trading performance are often conflicting objectives, and 
a model that ranks highly on classification metrics may not translate to a profitable trading strategy. 
Evaluating one as a proxy for the other produces systematically misleading conclusions~\cite{gu2020, 
Fieberg2023}.

To answer the proposed questions, we conduct a series of experiments. TDL models are incorporated
into a cross-sectional trading strategy on US market constituents. Hyperparameters are selected 
via cross-regime Bayesian optimisation and strategies are evaluated on held-out 2025 data, with 
trading performance as the primary criterion. A robustness analysis on the top performing 
strategy provides a view of sensitivity to regime and input shifts, along with 
feature attributions calculated through SHAP values to determine the contribution 
of each source category.
The key contributions are as follows:

\begin{itemize}[nosep]
    \item The Hybrid ensemble maintains performance across 
    statistically distinct market regimes and realistic levels 
    of input perturbation, providing evidence that 
    cross-regime Bayesian optimisation leads to generalisation beyond the 
    conditions under which hyperparameters were estimated. This 
    approach to regime-robust tuning is not specific to equities 
    or finance, and could extend to any asset class or 
    task where verifiable regimes exist.

    \item TDL architectures do not provide 
    consistent individual improvements over gradient-boosted 
    trees in cross-sectional equity prediction. Their value 
    emerges within an ensemble, with
    XGBoost and TabNet producing an ensemble with 
    significant OOS alpha.

    \item Alternative data contributes a supplementary 
    role. Attribution varies by model architecture and 
    contributes more to short signal generation than long in 
    the ensemble constituents.
\end{itemize}

\begin{figure}[H]
    \centering
    \includegraphics[width=0.8\textwidth]{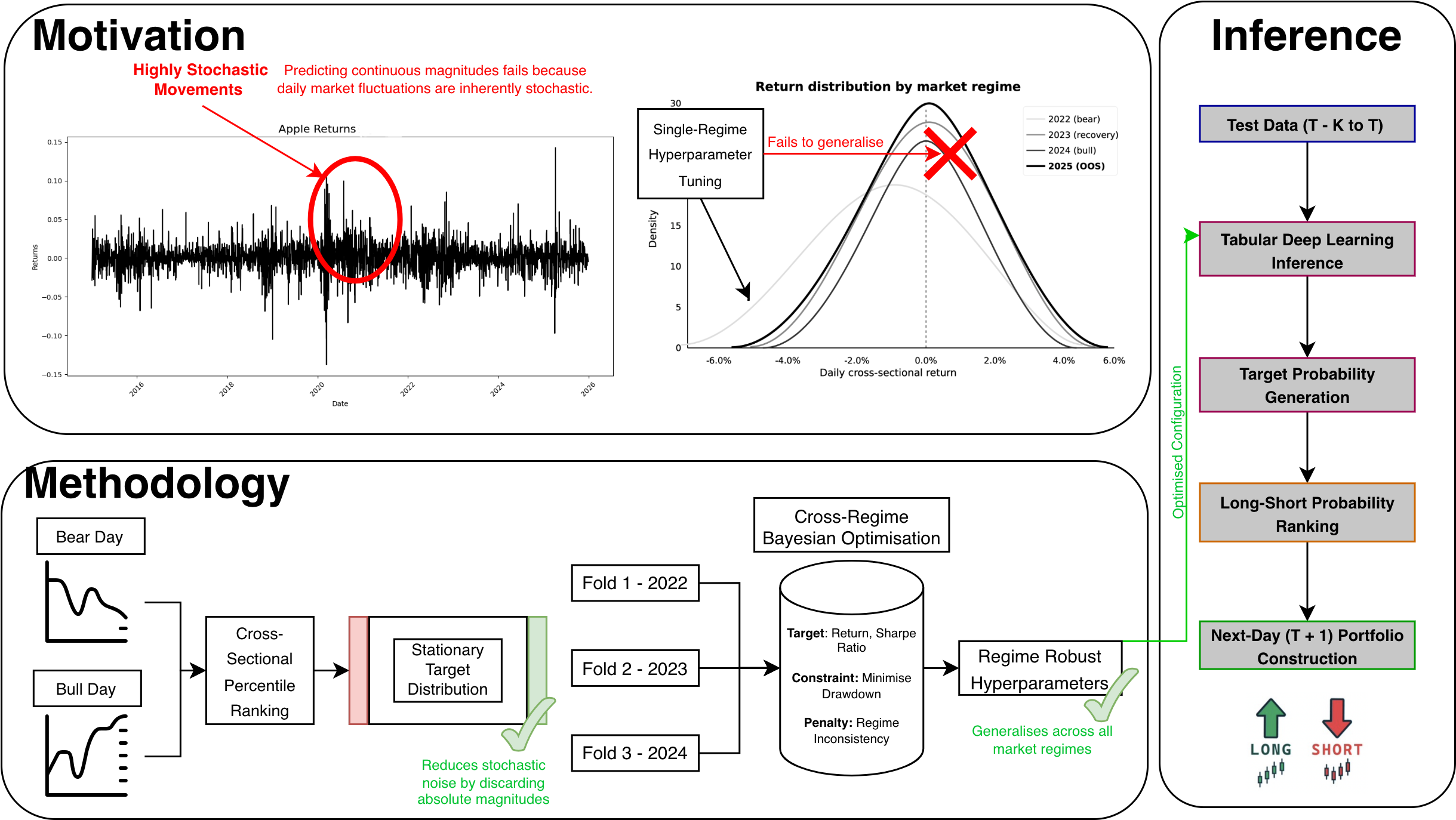}
    \vspace{-0.5em}
    \caption{Overview of motivation, methodology, and inference framework}\label{fig:contri_framework}
    \footnotesize
    (Top-Left) Single-regime hyperparameter tuning fails to
    capture highly stochastic, regime-dependent return distributions. (Bottom-Left) 
    Cross-sectional percentile ranking maps these heterogeneous 
    environments into a stationary target distribution. Cross-regime 
    Bayesian optimisation selects hyperparameters that generalise 
    across them. (Right) The daily execution pipeline processes historical data via TDL models to generate 
    target probabilities for OOS long-short portfolio construction
\end{figure}

Figure~\ref{fig:contri_framework} provides an overview of the 
complete framework developed to address these questions.

\section{Literature Review}

\subsection{Traditional Methods}\label{sec:trad_methods}

Financial modelling began with continuous-time methods such as 
geometric Brownian motion~\cite{Samuelson1965} and the Black-Scholes model~\cite{BlackScholes1973}, which provide the 
first mathematical approach to asset price forecasting. Reliance on assumptions 
of constant volatility and log-normality has limited the accuracy 
of their return series predictions~\cite{BlackScholes1973}. ARIMA and GARCH 
improved forecasting by capturing autocorrelation and 
volatility clustering in univariate return series~\cite{Box1976, Bollerslev1986}. 
Both are still useful benchmarks for univariate forecasting, but they were 
designed for individual temporal sequences, reducing their relevance for 
heterogeneous cross-sectional data. 
In current research, this leaves them as baseline models and highlights the need for more
flexible methods that are limited by fewer assumptions~\cite{gu2020,Fieberg2023,Kumbure2022,Ma2021}.

Classical ML offers this structural flexibility by addressing
the linearity constraints of statistical forecasting and allowing for multivariate analysis. 
In contrast to traditional time-series models, ML is able to capture 
interactions among predictors that linear counterparts cannot. Research finds that Support Vector 
Machines~\cite{HuangWei2005Fsmm} and tree-based ensembles such as Random 
Forests~\cite{KhaidemLuckyson2016Ptdo,LohrmannChristoph2019CoiS}, and 
XGBoost~\cite{BasakSuryoday2019} improve forecasting over 
statistical baselines on financial applications. 
However, these models are limited by treating stocks as independent observations. 
As implied by Gu et al.~\cite{gu2020}, return predictability is 
driven by relative characteristics between equities, meaning that the most economically 
valuable component of the signal is discarded. Their 
predictive performance may also be dependent on manual feature 
engineering, requiring substantial domain expertise to construct 
informative inputs~\cite{Li2025}.

Deep learning reduced this need for feature engineering by learning complex feature 
representations from raw data~\cite{Zhang2025deep, Babiak2026}. Multi-Layer 
Perceptrons (MLP) applied to asset pricing have improved OOS return 
prediction over linear benchmarks~\cite{gu2020}, but lack mechanisms for 
modelling temporal dependence or handling irregular feature interactions. Recurrent 
neural networks, including their variants, address the temporal limitation by 
modelling sequential dependence across observations~\cite{Fischer2018} 
and Transformer-based models extend this to longer-range dependencies through 
self-attention~\cite{GiantsidiSofia2025Dlff}. Both sequential approaches are 
designed for homogeneous time-series data, making them poorly suited to the 
heterogeneous tabular feature sets that define cross-sectional equity panel data, 
where inputs cover continuous values, sentiment scores, and macroeconomic indicators 
with no inherent sequential structure.

\subsection{Tabular Deep Learning}\label{sec:tabular_dl}

Grinsztajn et al.~\cite{Grinsztajn2022} identify two structural reasons why 
standard neural networks struggle on tabular data.
Neural networks are biased toward smooth functions, while tabular target functions 
tend to be irregular and piecewise, favouring tree-based splits over gradient 
descent. MLPs are also disproportionately sensitive to uninformative features 
that are common in tabular datasets.
Borisov et al.~\cite{Borisov2024} reinforce this, outlining that standard MLPs 
have limited ability to identify features that dominate predictions 
or capture irregular interaction patterns, properties that tree-based methods 
handle naturally through their splitting structure. MLPs often 
require substantially more tuning and data to match what tree-based methods 
achieve~\cite{Grinsztajn2022, ShwartzZiv2022}, helping explain the consistent 
strength of gradient-boosted trees on benchmark tabular tasks.

TabNet~\cite{Arik2021} approaches tabular learning through sequential attention for 
instance-wise feature selection. Its sparsemax layers selectively focus on the most 
relevant features at each prediction step, designed to address the sensitivity to 
uninformative features that limits MLP performance, as well as mimicking 
the node-splitting behaviour of decision trees. 
The Feature Tokeniser-Transformer (FT-Transformer)~\cite{Gorishni2021} takes a 
different approach, representing each feature as a distinct token and applying 
multi-head self-attention across the feature set. By treating each feature as a distinct 
token, attention computes how strongly 
features relate to one another separately for each input. 
Related models such as TabTransformer~\cite{HuangXin2020} 
and SAINT~\cite{Somepalli2021} extend the architectural space further, but require 
categorical feature structure or substantial label-scarce pretraining that are 
not suited to the continuous tabular inputs present in equity panel data. 
TabNet and the FT-Transformer are therefore more architecturally appropriate 
for this research.

Performance across TDL architectures is sensitive to dataset 
structure and tuning. Benchmark studies demonstrate these models frequently 
outperform standard MLPs but rarely surpass gradient-boosted trees when 
hyperparameters are not carefully 
selected~\cite{Borisov2024,ShwartzZiv2022,Grinsztajn2022}. In financial 
applications, this sensitivity is exacerbated by regime changes that make robust 
generalisation more difficult. A configuration 
selected on a single validation period may be well-suited to the return 
distribution of that regime but fail to generalise when market conditions 
shift, yet existing evaluations of TDL models rely on static benchmarks with stable feature 
distributions~\cite{Borisov2024}. No existing study evaluates 
TabNet or FT-Transformer within a cross-sectional equity trading framework 
that explicitly targets regime robustness during hyperparameter selection.

\subsection{Data Sources}\label{sec:data_sources}

The tabular inputs within equity panel data contain 
many different source types, each capturing dimensions 
of stock and investor behaviour. Kumbure et al.~\cite{Kumbure2022} 
reviewed 138 financial ML studies from 2000 
to 2019. They identified 2173 unique variables covering technical indicators, 
macroeconomic factors, fundamental indicators, and alternative sources. 
Technical indicators are the focus of many studies despite criticism that their lagging 
behaviour reflects past price action~\cite{Kumbure2022}. 
Recent evidence confirms that alternative data sources are useful 
in investment practice, as big data inputs improve earnings forecasts 
and stock price prediction~\cite{ChiFeng2025, SunYunchuan2024}, substantiating 
their inclusion.

Firm-level news sentiment is a widely reviewed alternative data source 
for equity prediction as it captures information 
separate to historical return dynamics.
Atkins et al.~\cite{Atkins2018} find that information extracted from news 
carries predictive content for equity and index volatility, outperforming 
price-based prediction of directional movements, with Allen et al.~\cite{Allen2019} 
extending this to prices. More recently, 
Gambarelli and Muzzioli~\cite{Gambarelli2025} discuss that firm-level sentiment 
indicators are significantly priced in European stock returns, extending 
this evidence beyond the US market. The current literature, 
however, uses aggregate or index-level sentiment 
measures. Their relevance for cross-sectional prediction is limited, since 
the signal must differentiate between individual stocks on a given day.

Search volume data from Google Trends provides a proxy for investor attention that is 
not captured by other sources. Preis et al.~\cite{Preis2013} and Fan et al.~\cite{Fan2021} show that changes 
in frequency for finance-related search terms correlate with, and can serve as 
early indicators of, market movements. Szczygielski et al.~\cite{Szczygielski2024} 
build on this by establishing that such search-based signals are systematically 
related to market uncertainty. This evidence remains 
predominantly market-level as stock-specific search behaviour 
that could support cross-sectional differentiation is scarce.
Chen et al.~\cite{Chen2022} complicate this picture, showing 
that high-attention stocks experience short-term price pressure 
and that broad market-level indices may capture this signal 
despite their aggregation. This contrasts 
the granularity limitation raised above for sentiment 
measures, suggesting the value of aggregation may depend on the data source.
Macroeconomic indicators provide a further view of the broader context. 
Variables such as inflation, monetary policy stance, and yield curve 
dynamics contain information separate to firm-specific signals, 
capturing conditions shared across all stocks simultaneously~\cite{Kumbure2022}.

The construction of the dataset introduces risks that the 
literature has attempted to address. Survivorship bias inflates expected 
returns by excluding stocks delisted or removed from indices during the sample 
period~\cite{Brown1992}. Look-ahead bias arises from using data 
unavailable at prediction time. Fundamental data released 
with reporting lags and alternative sources with non-standard publication 
schedules need particular care to ensure only data available at 
the time are provided to the model~\cite{Yae2024}. The relative 
contribution of these source types remains unquantified 
across long and short signal directions in the cross-sectional equity 
setting, since feature attribution has not been applied separately by 
direction.

\subsection{Performance Evaluation Under Regime Instability}\label{sec:regime_eval}

Predictive accuracy and trading performance are not equivalent objectives in 
cross-sectional trading strategies. Classification accuracy weights every prediction 
equally regardless of the size or direction of the resulting return. Therefore, a model 
can achieve strong accuracy while its errors are concentrated in the highest-magnitude moves, 
producing poor trading performance independent of transaction costs. 
Applying realistic execution assumptions can magnify this, since 
apparent gains in returns can disappear entirely once transaction costs are 
applied~\cite{lopezdeprado2018, Ma2021}. Fieberg et al.~\cite{Fieberg2023} confirm 
this, showing that models ranking highest on accuracy do not 
consistently produce the strongest long-short portfolio returns. Portfolio-level 
evaluation is therefore a necessary condition for meaningful model 
comparison~\cite{gu2020, Xu2024}, but alone it does not guarantee reliable 
model selection, since a strategy validated on one market period may still 
fail once conditions shift.

Return distributions shift across market conditions, and a model tuned to one 
regime will often fail in another as the underlying signal changes~\cite{Timmermann2008, Ang2012}. 
One method has been to use regime detection, identifying market states through models 
such as Hidden Markov Models and switching model parameters or training sets accordingly. 
Pagliaro~\cite{Pagliaro2026} reports
economically meaningful risk-adjusted performance in cross-sectional equity 
settings using a regime-aware LightGBM framework. 
The limitation is that regime detection introduces its own sources of error. Misclassification 
of the current regime propagates into model selection, and the additional tuning 
required for each detected state increases the risk of overfitting to historical regimes
that may not recur.

Regime robustness can instead be built into the optimisation process itself, avoiding 
the need to detect regimes at inference time. Hyperparameters can be selected to 
perform well across multiple distinct environments simultaneously, 
without conditioning on a detected state, reducing sensitivity to any single regime's 
return distribution. Wong and Barahona~\cite{Wong2023} treat distribution shift as a 
structural property of financial data supporting treating regime 
heterogeneity as a constraint within optimisation itself.
Signal decay compounds this, as published predictive signals erode once 
market participants exploit them~\cite{Mclean2016}, so a configuration 
validated on a single historical regime is unlikely to hold going 
forward.

\section{Methodology}\label{sec:method}

\subsection{Data}

\subsubsection{Dataset and Preprocessing}\label{sec:preprocessing}

\begin{figure}[H]
    \centering
    \begin{center}
\begin{tikzpicture}[
    scale=0.75, transform shape,
    node distance=0.6cm,
    every node/.style={font=\fontsize{9}{11}\selectfont\sffamily},
    sourcebox/.style={
        rectangle, draw=black!70, thick, rounded corners=4pt,
        fill=black!4, minimum width=4.5cm, minimum height=0.5cm,
        align=center
    },
    mainbox/.style={
        rectangle, draw=black!70, thick, rounded corners=4pt,
        fill=black!6, minimum width=15cm, minimum height=1cm,
        align=center
    },
    targetbox/.style={
        rectangle, draw=black!70, thick, rounded corners=4pt,
        fill=black!10, minimum width=15cm, minimum height=1cm,
        align=center
    },
    signalbox/.style={
        rectangle, draw=black!70, thick, rounded corners=3pt,
        fill=black!4, minimum width=4cm, minimum height=1.1cm,
        align=center
    },
    finalbox/.style={
        rectangle, draw=black!70, thick, rounded corners=4pt,
        fill=black!15, minimum width=15cm, minimum height=1cm,
        align=center
    },
    arrow/.style={-{Latex[length=2.5mm]}, thin, black!70},
]

\node[sourcebox] (fred) {
    \textbf{FRED API}
    };

\node[sourcebox, left=0.75cm of fred] (bloomberg) {
    \textbf{Bloomberg Terminal}
    };

\node[sourcebox, right=0.75cm of fred] (trends) {
    \textbf{Google Trends}
    };

\node[mainbox, below=0.4cm of fred] (integration) {
    \textbf{Data Integration \& Processing}\\[1pt]
    Temporal Alignment $\cdot$ Feature Lags $\cdot$
    VIF Removal
};

\draw[arrow] (bloomberg.south) -- ++(0,-0.2) -| (integration.north);
\draw[arrow] (fred.south) -- (integration.north);
\draw[arrow] (trends.south) -- ++(0,-0.2) -| (integration.north);

\node[mainbox, below=0.4cm of integration] (features) {
    \textbf{Feature Engineering}\\[1pt]
    Momentum $\cdot$ Technical indicators $\cdot$
    Sentiment $\cdot$ Liquidity $\cdot$ Macroeconomic
};

\draw[arrow] (integration) -- (features);

\node[targetbox, below=0.4cm of features] (target) {
    \textbf{Target Construction}\\[1pt]
    Cross-sectional Next Day Return Percentiles
};

\draw[arrow] (features) -- (target);

\node[signalbox, below=0.4cm of target] (hold) {
    \textbf{Hold}\\middle --- 80\%
};

\node[signalbox, left=1.5cm of hold] (long) {
    \textbf{Long}\\top decile --- 10\%
};

\node[signalbox, right=1.5cm of hold] (short) {
    \textbf{Short}\\bottom decile --- 10\%
};

\draw[arrow] (target.south) -- ++(0,-0.2) -| (long.north);
\draw[arrow] (target.south) -- (hold.north);
\draw[arrow] (target.south) -- ++(0,-0.2) -| (short.north);

\node[finalbox, below=0.4cm of hold] (final) {
    \textbf{Final Dataset} \\[1pt]
    $\sim$300 S\&P 500 Constituents $\cdot$
    2015--2025 $\cdot$ Daily Stock-Day Observations
};

\draw[arrow] (long.south) -- ++(0,-0.2) -| (final.north);
\draw[arrow] (hold.south) -- (final.north);
\draw[arrow] (short.south) -- ++(0,-0.2) -| (final.north);

\end{tikzpicture}
\end{center}
    \caption{Dataset Construction}\label{fig:data_pipeline}
    \footnotesize
    Traditional market data, company fundamentals, and news sentiment from Bloomberg are integrated 
    with Google Trends and FRED macroeconomic indicators. The processed data generates next-day 
    cross-sectional return percentiles, establishing outer-decile (10\%) long and short target 
    classes for approximately 300 S\&P 500 constituents
\end{figure}
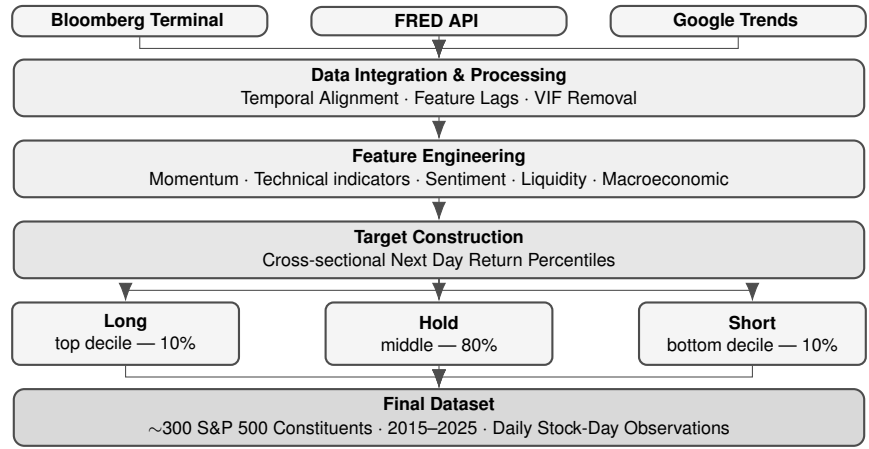

The collection universe contained 300 large-cap S\&P 500 constituents from
January 2015 to December 2025, as illustrated in Figure~\ref{fig:data_pipeline}. Restricting 
the universe to large-cap constituents ensured sufficient liquidity to make backtested 
transaction cost assumptions realistic. In addition to price and company data, macroeconomic indicators from FRED~\cite{fred2026} 
capture broader market conditions. Firm-level news sentiment~\cite{bloomberg2026} and Google Trends 
indices~\cite{googletrends2026} provide signals of information content 
orthogonal to price dynamics and market-wide behavioural 
attention~\cite{Atkins2018, Allen2019, Chen2022}.

Data cleaning was conducted to ensure the integrity of the dataset. Companies with missing price observations were dropped, as imputing prices could introduce skew into return 
series, with fundamental values being forward-filled within each ticker to ensure the most recently reported figure 
was available. Days where no sentiment score was reported were assigned a sentiment score of 
zero, the assumption being neutrality where there is no coverage. All weekend news is aggregated into the
following trading day to ensure the inclusion of its signal. No outlier removal was applied to return observations, 
as extreme values in financial data carry signal and their removal could discard tail events most relevant 
to a long-short strategy. 

To prevent look-ahead bias, monthly macroeconomic indicators were shifted forward by 21 trading days, 
quarterly fundamentals by 63 days, weekly jobless claims by 5 days, and news and search volume 
data by one trading day, ensuring only information available at the point of prediction 
was used. Google Trends required additional handling, as each 
downloaded batch is independently normalised to a 0--100 scale, making consecutive batches 
incomparable. A chained normalisation approach corrected this by computing a median scaling 
factor from overlapping periods between consecutive windows to construct a continuous time series.

Finally, multicollinearity among features was assessed using the Variance Inflation Factor (VIF), 
with a threshold of 10 applied as a commonly used rule-of-thumb for feature removal~\cite{Obrien2007}. 
Features exceeding this threshold were removed when redundant, although a small number of high-VIF 
features were retained where domain relevance outweighed collinearity concerns, since VIF 
thresholds are context-dependent and should not be treated as absolute decision rules~\cite{Obrien2007, Kalnins2025}.

\subsubsection{Feature Engineering}

The collected features cover four categories: technical indicators, 
company fundamentals, macroeconomic factors and alternative data, reflecting the categories 
identified by Kumbure et al.~\cite{Kumbure2022}. Raw features are insufficient for 
cross-sectional prediction, as careful transformation of inputs can improve 
OOS return predictability~\cite{Li2025}. Where possible, features were 
therefore constructed in relative terms. Return ranks, 
momentum, oscillators, and volume measures are expressed relative to either the cross-sectional 
distribution or each stock's own recent history, since absolute values are not 
informative for a prediction task whose objective is to rank stocks relative to 
one another. 

Alternative data sources required additional engineering beyond standard 
transformation due to the complexity of their relationships with returns. 
News sentiment was supplemented with a binary indicator flagging extreme 
observations, capturing potential non-linearity in the relationship between 
sentiment magnitude and returns that a linear feature would not represent. 
Equity market volatility was 
captured through a rolling quantile-based regime indicator, 
as the predictive content of volatility for cross-sectional 
returns derives from whether conditions are elevated relative to recent history.

\footnotesize
\begin{equation}
\begin{minipage}{0.9\linewidth}
\centering
$r_{t+1}^{(i)} = \ln\!\left(\dfrac{P_{t+1}^{(i)}}{P_t^{(i)}}\right)$
\hfill
$\text{Rank}_{t+1}^{(i)} = \operatorname{PercentileRank}\!\left(r_{t+1}^{(i)}\right)$
\hfill
$y_t^{(i)} = \begin{cases} 2 & \text{if } \text{Rank}_{t+1}^{(i)} > 0.90, \\ 0 & \text{if } \text{Rank}_{t+1}^{(i)} < 0.10, \\ 1 & \text{otherwise.} \end{cases}$
\end{minipage}
\label{eq:target_formulation}
\end{equation}

\normalsize

The target variable is defined in Equation~\eqref{eq:target_formulation}. The 
prediction task was conducted as a cross-sectional classification problem, 
since predicting continuous returns is problematic given 
their high noise, heteroskedasticity, and non-stationarity. Next-day returns were 
ranked into deciles across the universe daily. The top and bottom deciles 
assigned long and short labels, respectively. The outer-decile threshold aligns with 
established cross-sectional equity ML studies~\cite{gu2020, Fieberg2023}, 
concentrating positions in the highest-conviction signals and managing the class 
imbalance caused by using narrow threshold boundaries. UMAP projections of the 
training feature space, presented in Appendix~\ref{app:umap}, show that long and 
short labels are intermixed without a clear structure, as well as structural 
groupings that reveal regime-related heterogeneity that is verified formally via KS 
tests in Section~\ref{sec:setup}.

\subsection{Experimental Setup}\label{sec:setup}

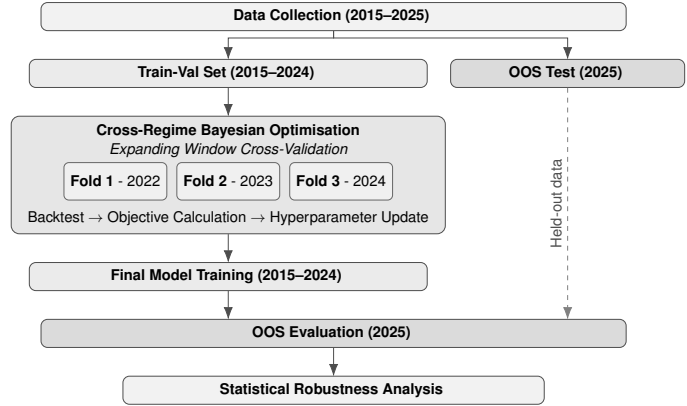
\begin{figure}[H]
    \centering
    \resizebox{0.55\textwidth}{!}{
        \begin{tikzpicture}[
    node distance=1.2cm and 0cm,
    every node/.style={font=\fontsize{9}{11}\selectfont\sffamily},
    box/.style={rectangle, draw=black!70, thin, rounded corners=4pt,
                align=center, minimum height=0.6cm},
    foldbox/.style={rectangle, draw=black!60, thin, rounded corners=3pt,
                    align=center, fill=black!6, minimum height=0.8cm, minimum width=1.9cm},
    container/.style={rectangle, draw=black!70, thin, rounded corners=5pt, 
                      fill=black!10, inner sep=4pt},
    arrow/.style={-{Latex[length=2.5mm]}, thin, black!70},
    dasharrow/.style={-{Latex[length=2.5mm]}, thin, black!50, dashed},
]

\node[box, fill=black!5, minimum width=12.5cm] (data) {
    \textbf{Data Collection (2015--2025)}
};

\node[box, fill=black!8, minimum width=8.5cm, below=0.6cm of data.south, xshift=-2.25cm] (trainval) {
    \textbf{Train-Val Set (2015--2024)}
};

\node[box, fill=black!14, minimum width=5cm, below=0.6cm of data.south, xshift=5cm] (ooslock) {
    \textbf{OOS Test (2025)}
};

\draw[arrow] (data.south) -- ++(0,-0.15) -| (trainval.north);
\draw[arrow] (data.south) -- ++(0,-0.15) -| (ooslock.north);

\node[container, below=0.6cm of trainval, minimum width=8.5cm] (optuna) {
    \begin{tabular}{c}
        \textbf{Cross-Regime Bayesian Optimisation} \\
        \textit{Expanding Window Cross-Validation} \\[2pt]
        \begin{tikzpicture}
            \node[foldbox] (f1) {\textbf{Fold 1} - 2022};
            \node[foldbox, right=0.2cm of f1] (f2) {\textbf{Fold 2} - 2023};
            \node[foldbox, right=0.2cm of f2] (f3) {\textbf{Fold 3} - 2024};
        \end{tikzpicture} \\ [2pt]
        Backtest $\rightarrow$ Objective Calculation $\rightarrow$ Hyperparameter Update
    \end{tabular}
};
\draw[arrow] (trainval.south) -- (optuna.north);

\node[box, fill=black!8, minimum width=8.5cm, below=0.6cm of optuna] (finalmodel) {
    \textbf{Final Model Training (2015--2024)}
};
\draw[arrow] (optuna.south) -- (finalmodel.north);

\node[box, fill=black!14, minimum width=12.5cm, below=0.6cm of finalmodel, xshift=2.25cm] (ooseval) {
    \textbf{OOS Evaluation (2025)}
};

\draw[arrow] (finalmodel.south) -- (finalmodel.south |- ooseval.north);
\draw[dasharrow] (ooslock.south) -- (ooslock.south |- ooseval.north)
    node[midway, left, rotate=90, anchor=south, font=\footnotesize\sffamily, text=black!70] {Held-out data};

\node[box, fill=black!5, minimum width=9cm, below=0.6cm of ooseval] (stats) {
    \textbf{Statistical Robustness Analysis}
};
\draw[arrow] (ooseval.south) -- (stats.north);

\end{tikzpicture}
    }
    \caption{Modelling Methods}\label{fig:method}
    \footnotesize
    Data is split into a training and validation set and an out-of-sample (OOS) test set. 
    Cross-regime Bayesian optimisation utilises an expanding window cross-validation across 
    distinct market regimes before final model training and OOS evaluation
\end{figure}

\vspace{-0.5em}

\subsubsection{Data Partitioning and Regime Validation}\label{sec:partitioning}

The full dataset was divided into training and validation sets covering 2015 to 
2024 and an OOS test set comprising 2025, as illustrated 
in Figure~\ref{fig:method}. The 2025 period was excluded from model 
development and hyperparameter selection to provide an uninfluenced estimate of 
real-world performance.

Three validation folds were defined within the training period, corresponding to 
2022, 2023, and 2024, constructed using an expanding-window approach. 
Each fold trains on all data from 2015 through the preceding year and evaluates 
on the target year. This design was chosen instead of a rolling window, which discards older 
observations, since retaining the full historical training set reflects 
production conditions where all available data would be utilised. Standard k-fold 
cross-validation was excluded as random shuffling of time-series observations 
would introduce look-ahead bias, where future observations appear in the training 
set of earlier folds~\cite{lopezdeprado2018}. 
A one-day purge was applied at 
each fold boundary, removing the final training observation whose label window 
extended into the validation period. Pairwise two-sample Kolmogorov-Smirnov (KS) tests were applied 
to daily return and 5-day realised volatility distributions 
across all fold pairs to verify that each period is a 
different market regime. Distributional differences were found to be significant 
across all regimes ($p < 0.001$); full results are reported in 
Appendix~\ref{app:ks}. A further KS test comparing the full training distribution 
against 2025 confirmed a meaningful distributional shift in the OOS period, 
reinforcing the need for multi-regime validation. Because this shift is 
present, any strong OOS performance that 
follows would provide evidence that the cross-regime framework enables 
model generalisation across market shifts.

A RobustScaler was fitted on the training portion of each fold and standardises 
all input features, with its learned parameters then applied to the validation 
sets to prevent data leakage. RobustScaler was implemented instead of standard normalisation 
due to the high skew and kurtosis of financial return distributions, 
as it scales features using the interquartile range.
The three-class target distribution was 
highly imbalanced by construction, since the hold class contains approximately 80\% 
of observations. Therefore, tree-based and linear models utilise balanced class weights 
and deep learning models use a class-weighted cross-entropy loss 
to penalise minority class errors proportionally.

\subsubsection{Backtesting Framework and Portfolio Construction}\label{sec:backtesting}

During validation and final evaluation, each model produces a three-class probability 
vector for every stock-day observation, the full inference pipeline is illustrated 
in Figure~\ref{fig:contri_framework}. Daily trading signals were generated by 
ranking stocks cross-sectionally according to their predicted long and short class 
probabilities separately, with the top $n$ long stocks assigned long positions and the 
top $n$ short stocks assigned short positions. Separating the long and short 
rankings ensures each signal direction is evaluated on its own cross-sectional 
distribution, so a stock qualifies for each trading book on the strength of 
its probability relative to other stocks. The \$10 million initial capital was split evenly 
across the long and short books and equally across positions within each side. This
gives a dollar-neutral, unlevered portfolio with $100\%$ gross exposure and 
avoids amplifying the influence of any single holding. The portfolio was 
rebalanced daily and trades are executed at market-on-close prices. Features requiring 
same-day price information were constructed from a pre-close snapshot and orders are
submitted ahead of the market-on-close cut-off. The closing price was used as an execution 
proxy, on the assumption that the difference between the pre-close and closing price 
is negligible. A fixed cost of 2.2 basis points per trade accounts for transaction costs and 
slippage, charged per leg on entry and exit, including both legs of a same-day 
flip. Hagstromer~\cite{Hagstromer2021} reports a mean effective spread of 2.84 
basis points for S\&P 500 constituents, implying a one-way execution cost of 1.42 
basis points. The assumption applied here therefore exceeds the spread component, 
leaving margin for commissions and fees. Strategy performance was benchmarked 
against a passive S\&P 500 buy-and-hold strategy. This same portfolio construction 
procedure was applied during hyperparameter optimisation and final OOS 
evaluation across fixed random seeds to support reproducibility.

\subsubsection{Bayesian Hyperparameter Optimisation}

Hyperparameters for all models are selected using Bayesian optimisation via 
the Optuna framework~\cite{Akiba2019}, utilising a Tree-structured Parzen Estimator (TPE)~\cite{Bergstra2011} 
sampler with 30 trials per model. The objective function was designed to 
target trading performance across all three validation regimes simultaneously, in 
place of optimising for predictive accuracy in a single validation fold.
This design targets the gap between classification performance and realised trading outcomes discussed by 
López de Prado~\cite{lopezdeprado2018} and Gu et al.~\cite{gu2020}. For each Optuna trial, 
the model was trained and backtested, details shown in Section~\ref{sec:backtesting}, on all three 
validation folds, and a composite score was computed as shown in Equation~\ref{eq:optuna_fitness}.

\footnotesize
\begin{equation}
    \begin{gathered}
    \text{Score} = \underbrace{ 0.4 \left( \frac{\bar{r}}{0.15} \right) + 0.4 \left( \frac{\bar{s}}{1.5} \right) - 0.2 \left( \frac{\bar{d}}{0.10} \right) }_{\text{Base Objective}} - \underbrace{ \left( P_r + P_s + P_d + P_{\text{floor}} \right) }_{\text{Penalties}} \\[0.5em]
    P_r = 1.5 \left( \frac{\max_i | \min(0, r_i) |}{0.15} \right)^2 \qquad P_s = 1.5 \left( \frac{\max_i | \min(0, s_i) |}{1.5} \right)^2 \\
    P_d = 1.0 \left( \frac{\max_i | \max(0, d_i - 0.15) |}{0.15} \right)^2 \qquad P_{\text{floor}} = \begin{cases} 10.0 & \text{if } \exists\, i : r_i < -0.20 \\ 0.0 & \text{otherwise} \end{cases} \\[0.5em]
    \footnotesize\text{where } \bar{r},\,\bar{s},\,\bar{d} \text{ denote mean return, Sharpe ratio, and drawdown across all regimes.}
    \end{gathered}
    \label{eq:optuna_fitness}
\end{equation}

\normalsize
The base objective combines three trading metrics. Mean annualised return and 
Sharpe ratio each receive 40\% weight as the performance objectives, with maximum 
drawdown at 20\% as a risk management constraint. Each metric is divided by a target 
value before weighting, 0.15 for return, 1.5 for Sharpe, and 0.10 for drawdown, 
so that a metric meeting its target contributes a ratio of one to the score. 
Without this, the differing magnitudes of the three metrics would distort the 
intended emphasis of the weights. All weights were fixed before experimentation 
began, since tuning them alongside the model would make the objective function 
itself a source of overfitting.

The targets are calibrated against S\&P~500 benchmarks observed over the 
training period. A return target of 15\% exceeds the mean annualised index 
return of 14.3\%; the Sharpe target of 1.5 exceeds the benchmark Sharpe of 0.93; 
and the drawdown target of 10\% improves upon the mean annual drawdown of 13.1\%. 
In each case the threshold demands outperformance of the passive benchmark.

The main risk in cross-regime optimisation is that the optimiser identifies 
configurations that perform well on average by excelling in one regime and 
collapsing in another. Quadratic penalties are applied to prevent this. 
Return and Sharpe penalties are triggered by any negative fold, 
and a drawdown penalty triggered when any fold exceeds 15\%. 
Configurations with drawdown between 10\% and 15\% are discouraged through the 
base score but not eliminated, preserving flexibility in the hyperparameter search. 
The quadratic penalty means a drawdown of 20\% incurs four times the 
cost of one at 10\%, making severe violations disproportionately 
expensive. A hard floor penalty of 10.0 eliminates any 
configuration returning below $-20\%$ in any single regime. Portfolio size $n$ was 
tuned jointly with model hyperparameters within a search range of 5 to 8.

\subsection{Model Architectures}

Five models were implemented covering linear, tree-based, and deep learning 
approaches, with Table~\ref{tab:model_config} summarising the architecture 
and training configuration for each. All hyperparameters were selected via 
the Bayesian optimisation framework described in Section~\ref{sec:setup} 
unless otherwise stated. Search bounds are informed by dataset scale, 
computational constraints, and established ranges in the literature.

\subsubsection{Baseline Models}

Three baseline models were implemented to establish performance benchmarks. 
Logistic Regression (LR) served as a linear baseline, providing a lower bound on 
the complexity required to generate predictive trading signals. 
Gradient-boosted trees remain state of the art on tabular 
classification tasks~\cite{Grinsztajn2022}, therefore XGBoost was included 
as a tree-based baseline and was expected to provide the most 
competitive non-deep-learning comparison. An MLP was 
included as a deep learning baseline to isolate the contribution of 
architecture-specific inductive biases in TabNet and FT-Transformer from 
the general benefit of deep learning over tree-based methods.

\subsubsection{Models Under Evaluation}\label{sec:ensemble}

TabNet~\cite{Arik2021} and the FT-Transformer~\cite{Gorishni2021} were 
selected because each addresses a different MLP limitation identified in 
Section~\ref{sec:tabular_dl}. TabNet's sequential attention performs 
instance-wise feature selection, targeting the sensitivity to uninformative 
features, whereas the FT-Transformer's 
per-feature tokenisation and self-attention allow feature interactions to 
be computed dynamically for each input. Including the MLP 
alongside both isolates if these mechanisms deliver gains beyond 
generic deep learning capacity.

A hybrid ensemble was constructed by combining the predictions of XGBoost 
and TabNet via rank aggregation, selected as the optimal combination from 
all ensembles of the models. Ensemble selection involved removing any 
combination producing negative returns in any validation regime and then 
surviving combinations are ranked using the composite scoring function described 
in Equation~\ref{eq:optuna_fitness} applied across the three validation 
regimes. On each trading day, each model's long and short signal 
probabilities were ranked, and the resulting ranks were averaged across 
models before applying the top-$n$ selection logic. This approach is more 
robust to differences in probability calibration between models than 
averaging raw probabilities, ensuring that a poorly calibrated model cannot 
dominate the signal. The ensemble portfolio size $n$ was set as the integer 
mean of the individual model portfolio sizes.

\begin{table}[h]
\centering
\footnotesize
\setlength{\tabcolsep}{4pt}
\caption{Model Architecture and Training Configuration}
\label{tab:model_config}
\begin{tabular}{p{2.0cm}p{3.5cm}p{1.5cm}p{8.5cm}}
\toprule
\textbf{Model} & \textbf{Key Architecture} & 
\textbf{Early Stopping} & \textbf{Hyperparameter Search Space} \\
\midrule
\raggedright Logistic Regression & 
\raggedright Linear model & 
-- & 
Regularisation strength $C \in [1\text{e-}5, 1\text{e}4]$ \\
\addlinespace
\raggedright XGBoost & 
\raggedright Gradient boosted trees & 
30 rounds & 
Estimators $\in [1000, 1300]$, learning rate $\in [0.005, 0.1]$, max depth $\in [3, 5]$, 
min child weight $\in [15, 50]$, $\gamma \in [0.5, 2.0]$, $\lambda \in [0.5, 3.0]$, 
$\alpha \in [1.0, 2.0]$, subsample $\in [0.7, 1.0]$, column sample $\in [0.5, 1.0]$ \\
\addlinespace
\raggedright MLP & 
\raggedright Halving layers, LayerNorm, ReLU & 
10 epochs & 
Hidden dimension $\in \{128, 256, 512\}$, layers $\in [2, 3]$, dropout $\in [0.1, 0.3]$, 
learning rate $\in [1\text{e-}4, 5\text{e-}3]$, batch size $\in \{2048, 4096\}$, 
gradient clipping $\in [0.5, 2.0]$, weight decay $\in [1\text{e-}5, 1\text{e-}3]$ \\
\addlinespace
\raggedright TabNet & 
\raggedright Sequential attention, sparsemax, ghost batch normalisation & 
10 epochs & 
$n_d{=}n_a \in \{8, 16, 32\}$, decision steps $\in \{3, 4\}$, $\gamma \in [0.5, 3.0]$, 
learning rate $\in [5\text{e-}4, 1\text{e-}2]$, batch size $\in \{2048, 4096, 8192\}$, 
gradient clipping $\in [0.5, 1.5]$, weight decay $\in [5\text{e-}4, 1\text{e-}1]$ \\
\addlinespace
\raggedright FT-Transformer & 
\raggedright Feature tokenisation, multi-head self-attention, linear head & 
10 epochs & 
Attention blocks $\in \{2, 3\}$, attention heads $\in \{2, 4, 8\}$, 
embedding dimension $\in \{32, 64\}$, head size $\in \{32, 64\}$, 
attention dropout $\in [0.2, 0.5]$, feed-forward dropout $\in [0.2, 0.5]$, 
learning rate $\in [5\text{e-}4, 1\text{e-}2]$, batch size $\in \{2048, 4096, 8192\}$, 
gradient clipping $\in [0.5, 1.5]$, weight decay $\in [5\text{e-}4, 1\text{e-}1]$ \\
\bottomrule
\end{tabular}
\vspace{0.3em}
\begin{minipage}{15.5cm}
\footnotesize\raggedright
\vspace{0.5em}
All deep learning models trained with AdamW optimiser and ReduceLROnPlateau learning rate scheduler for 100 epochs with
early stopping on the validation set.
Class-weighted cross-entropy loss with label smoothing $= 0.1$ where applicable.
TabNet uses a fixed virtual batch size of 256 for ghost batch normalisation.
\end{minipage}
\end{table}

\vspace{-0.5em}

\subsection{Evaluation Framework}

\subsubsection{Out-of-Sample Performance and Statistical Significance Testing}

After hyperparameter selection, the final configuration was retrained on the full 2015--2024 period and evaluated 
on the held-out 2025 set using the portfolio construction procedure described in Section~\ref{sec:backtesting}. 
Final OOS evaluation provides the primary estimate of real-world strategy performance accompanied by 
classification metrics offering a diagnostic of signal quality.

Daily strategy returns were assessed for normality using the Shapiro-Wilk test. Non-Gaussian characteristics 
are observed, meaning a suite of non-parametric tests was applied to compare model 
performance, with results significant at $\alpha < 0.05$. The KS test was used to compare the return distributions of competing strategies, 
the Friedman test was used to assess differences across models, and the Wilcoxon Signed-Rank test was used to 
compare individual models against the S\&P 500. To further evaluate risk-adjusted performance, the Probabilistic 
Sharpe Ratio (PSR)~\cite{lopezdeprado2018} was computed against the S\&P 500's daily Sharpe ratio as the reference threshold, 
accounting for skewness and excess kurtosis in the return distribution. 
CAPM regression of daily strategy returns on S\&P 500 benchmark returns was performed to estimate annualised alpha, beta, and $R^2$, 
isolating the extent to which performance was attributable to stock selection skill. 
Pairwise cross-regressions between models were also conducted to examine the degree of similarity between their predictive 
outputs.

Classification-based metrics are reported 
alongside portfolio-level metrics. However, since the research 
objective is to evaluate trading performance, the main criteria for model comparison are annual 
return, Sharpe ratio, drawdown, and statistical significance relative 
to the benchmark.

\vspace{-1em}

\subsubsection{Robustness and Temporal Stability Analysis}

Two analyses test the regime robustness and input stability predicted by the 
hypothesis stated in Section~\ref{sec:introduction}. Aggregate OOS 
performance over a single year cannot separate a regime-robust 
configuration from one that happened to suit 2025, nor can it establish 
if the signal survives the input degradation that vendor inconsistencies 
or pipeline latency would introduce in production. 

Input stability was assessed by injecting Gaussian noise into the standardised 
test features at four levels ($\sigma \in \{0.05, 0.10, 0.20, 0.50\}$) across 20 
random seeds. It was applied proportionally to each feature's standard deviation so that 
$\sigma$ reflects a fraction of each feature's own variability. 
Jensen-Shannon divergence quantified the resulting shift in predicted probability 
distributions. Portfolio return, Sharpe ratio, and maximum drawdown were recorded 
at each level to assess practical trading impact. A null baseline replacing 
all features with Gaussian noise was included to 
determine if the model is learning signal or performance is due to chance.

Regime generalisation was assessed by dividing 2025 into four quarterly 
sub-periods. Pairwise KS tests confirmed each quarter of 2025 was a 
distinct market environment. Performance across quarters therefore 
reflects generalisation to conditions the model was not tuned on.
Long and short signal precision were then evaluated separately within each quarter to 
determine whether predictive performance holds across changing conditions or 
degrades over time.

\vspace{-1em}

\subsubsection{Interpretability and Operationalisation}\label{sec:interpretability}

Signals that drive the model predictions are often masked by the black
box outputs of many ML models. To highlight valuable signals, SHAP values are computed for all five 
models using KernelExplainer with a k-means background of 100 cluster 
centres derived from the 2015--2024 training distribution.
KernelExplainer was chosen over architecture-specific alternatives to ensure 
attributions are accurately comparable across models without introducing 
explainer-specific variance. Values were approximated over 5,000 training 
observations with 500 background samples per call, to balance computational costs.

Separate SHAP values were computed for both signal directions. This 
distinction matters because a model generating asymmetric signals may not rely 
on the same features; collapsing long and short attributions into 
a single importance ranking would obscure this. Attributions 
were aggregated by data source category following the categories outlined by 
Kumbure et al.~\cite{Kumbure2022}. This allows assessment of the relative 
contribution of technical, fundamental, macroeconomic, and alternative inputs to 
each signal direction. The hybrid ensemble is excluded from SHAP analysis 
as rank aggregation does not expose a differentiable output.

Static performance metrics resist interrogation of the conditions or reasons 
behind a strategy's performance.
To address this, an interactive research application was 
developed exposing backtest results, SHAP attributions, and input features 
through a REST API with a React frontend. Users can adjust position 
limits and examine model behaviour across regimes in real time. An AI research 
assistant grounded in the study's results enables natural language queries over 
model performance and feature drivers, operationalising the outputs in a form 
closer to practical deployment.

\subsection{Ethical Considerations}

This project involved no human participants, personal data, or live capital, 
and did not require ethics approval. Bloomberg Terminal data was accessed 
under an institutional licence and not redistributed. 
The reliance on licensed data constrains 
independent verification, mitigated in part by the use of publicly available 
FRED and Google Trends sources alongside it.

The main ethical risk in backtested trading research 
is the reporting of performance that could not have been achieved in practice, 
since inflated results influence how capital is allocated. The lag schedule, 
fold purging, and per-fold scaler fitting described in 
Section~\ref{sec:preprocessing} therefore serve to maintain integrity. 
ML-driven trading signals also have the potential 
to contribute to market instability if deployed at scale, a consideration 
returned to in Section~\ref{sec:deployment}.

\section{Results}
\subsection{Trading Performance and Comparison}

\vspace{-1em}

\begin{table}[H]
\centering
\caption{Model Performance Across All Models and Market Regimes}
\label{tab:trading_performance}
\begin{threeparttable}
\footnotesize
\renewcommand{\arraystretch}{1.1}
\setlength{\tabcolsep}{0pt}
\begin{tabular*}{\textwidth}{@{\extracolsep{\fill}} ll rrrr @{\hspace{1.5em}} rrrrr}
\toprule
& & \multicolumn{4}{c}{\textbf{ML Classification Metrics}} & \multicolumn{5}{c}{\textbf{Backtesting Performance}} \\
\cmidrule{3-6} \cmidrule{7-11}
\textbf{Year} & \textbf{Model} & \textbf{Acc} & \textbf{Pre} & \textbf{Rec} & \textbf{F1} & \textbf{Ret\%} & \textbf{SR} & \textbf{Sor} & \textbf{Cal} & \textbf{DD\%} \\
\midrule
\addlinespace[0.1cm]
\multirow{7}{*}{\textbf{2022}} 
 & S\&P 500  & -- & -- & -- & -- & -18.99 & -0.90 & -1.21 & -0.78 & -24.47 \\
 & LogReg         & 0.792 & 0.434 & \textbf{0.368} & \textbf{0.369} & -6.15 & -0.56 & -0.52 & -0.50 & -12.20 \\
 & XGBoost        & \textbf{0.799} & \textbf{0.440} & 0.357 & 0.350 & 9.02 & 0.36 & 0.76 & 0.56 & -16.01 \\
 & MLP            & 0.794 & 0.404 & 0.349 & 0.337 & 2.56 & 0.05 & 0.38 & 0.19 & -13.76 \\
 & TabNet         & 0.786 & 0.397 & 0.353 & 0.346 & 7.62 & 0.33 & 0.88 & 0.95 & \textbf{-8.04} \\
 & FT-Transformer & 0.794 & 0.423 & 0.357 & 0.351 & \textbf{16.00} & \textbf{0.75} & \textbf{1.63} & \textbf{1.59} & -10.09 \\
 & Hybrid         & 0.795 & 0.431 & 0.359 & 0.354 & 5.92 & 0.22 & 0.75 & 0.49 & -12.11 \\
\addlinespace[0.1cm]
\midrule
\addlinespace[0.1cm]
\multirow{7}{*}{\textbf{2023}} 
 & S\&P 500  & -- & -- & -- & -- & 26.00 & 1.58 & 3.05 & 2.61 & -9.97 \\
 & LogReg         & 0.787 & 0.424 & \textbf{0.364} & \textbf{0.363} & -13.48 & -1.26 & -1.37 & -0.80 & -16.87 \\
 & XGBoost        & \textbf{0.796} & \textbf{0.439} & 0.356 & 0.348 & \textbf{8.02} & \textbf{0.33} & \textbf{0.85} & \textbf{0.59} & -13.53 \\
 & MLP            & 0.792 & 0.415 & 0.351 & 0.340 & -0.55 & -0.22 & 0.05 & -0.04 & -12.62 \\
 & TabNet         & 0.787 & 0.418 & 0.359 & 0.354 & 2.14 & -0.06 & 0.36 & 0.23 & \textbf{-9.21} \\
 & FT-Transformer & 0.791 & 0.423 & 0.356 & 0.350 & 2.62 & 0.03 & 0.34 & 0.20 & -12.87 \\
 & Hybrid         & 0.793 & 0.437 & 0.360 & 0.355 & 3.61 & 0.08 & 0.47 & 0.28 & -12.88 \\
\addlinespace[0.1cm]
\midrule
\addlinespace[0.1cm]
\multirow{7}{*}{\textbf{2024}} 
 & S\&P 500  & -- & -- & -- & -- & 25.28 & 1.58 & 2.43 & 3.01 & -8.41 \\
 & LogReg         & 0.790 & 0.438 & \textbf{0.368} & \textbf{0.369} & -8.60 & -0.75 & -0.81 & -0.42 & -20.53 \\
 & XGBoost        & \textbf{0.797} & \textbf{0.444} & 0.357 & 0.349 & \textbf{29.07} & \textbf{1.22} & \textbf{2.04} & \textbf{2.42} & -12.03 \\
 & MLP            & 0.794 & 0.426 & 0.353 & 0.343 & 7.64 & 0.30 & 0.75 & 0.52 & -14.67 \\
 & TabNet         & 0.789 & 0.420 & 0.359 & 0.355 & 6.54 & 0.28 & 0.81 & 0.87 & \textbf{-7.52} \\
 & FT-Transformer & 0.793 & 0.435 & 0.359 & 0.354 & 2.28 & 0.01 & 0.29 & 0.13 & -17.29 \\
 & Hybrid         & 0.794 & 0.442 & 0.361 & 0.356 & 25.99 & 1.32 & 2.17 & 2.53 & -10.26 \\
\addlinespace[0.1cm]
\midrule
\addlinespace[0.1cm]
\multirow{7}{*}{\textbf{2025}} 
 & S\&P 500  & -- & -- & -- & -- & 18.01 & 0.77 & 1.22 & 0.96 & -18.76 \\
 & LogReg         & 0.789 & 0.431 & \textbf{0.366} & \textbf{0.365} & -5.69 & -0.63 & -0.62 & -0.37 & -15.23 \\
 & XGBoost        & 0.794 & 0.430 & 0.354 & 0.344 & 33.61 & 1.37 & 2.92 & 3.92 & -8.58 \\
 & MLP            & \textbf{0.795} & 0.436 & 0.355 & 0.346 & 17.34 & 0.77 & 1.45 & 1.40 & -12.35 \\
 & TabNet         & 0.789 & 0.423 & 0.360 & 0.355 & 9.30 & 0.44 & 1.15 & 0.61 & -15.24 \\
 & FT-Transformer & 0.792 & 0.430 & 0.358 & 0.351 & 3.08 & 0.06 & 0.46 & 0.22 & -14.20 \\
 & Hybrid         & 0.793 & \textbf{0.437} & 0.359 & 0.354 & \textbf{51.26} & \textbf{2.44} & \textbf{5.35} & \textbf{6.60} & \textbf{-7.76} \\
\addlinespace[0.1cm]
\bottomrule
\end{tabular*}

\vspace{0.4em}
\setlength{\tabcolsep}{0pt}
\begin{tabular*}{\textwidth}{@{\extracolsep{\fill}} l r @{\hspace{6em}} r @{\hspace{5em}} r}
\toprule
\multicolumn{4}{c}{\textbf{Per-Class Precision --- Out-of-Sample Period (2025)}} \\
\midrule
\textbf{Model} & \textbf{Long} & \textbf{Hold} & \textbf{Short} \\
\midrule
\addlinespace[0.1cm]
LogReg     & 0.233 & 0.821 & 0.240 \\
XGBoost    & 0.239 & 0.814 & 0.239 \\
MLP        & \textbf{0.247} & 0.815 & 0.247 \\
TabNet     & 0.223 & \textbf{0.818} & 0.229 \\
FT-Transformer  & 0.227 & 0.816 & 0.246 \\
Hybrid     & 0.240 & 0.817 & \textbf{0.254} \\
\addlinespace[0.1cm]
\bottomrule
\end{tabular*}
\begin{tablenotes}
\footnotesize
\item \textbf{Note}: 2025 is the out-of-sample period. Bold indicates the best model 
per period, excluding the benchmark. S\&P 500 ML metrics are not applicable (--). 
Sharpe ratios use excess returns 
with a fixed annualised risk-free rate of 3.6\%~\cite{fred2026}, applied identically to 
strategy and benchmark. Long and 
Short per-class precision exceed the 10\% random baseline, with 
the Hybrid achieving the highest Short precision.
\item \textbf{Abbreviations}: Acc = Accuracy, Pre = Precision, Rec = Recall, F1 = F1-Score, Ret\% = Total Return, SR = Sharpe, Sor = Sortino, Cal = Calmar, DD\% = Maximum Drawdown.
\end{tablenotes}
\end{threeparttable}
\end{table}

Table~\ref{tab:trading_performance} presents model performance across all four 
periods. The cross-validation years are included for completeness but are not 
independent evaluation, as they were used for hyperparameter selection. 
Aggregate classification metrics vary little across models as accuracy ranges
from 0.786 to 0.799 and F1 from 0.337 to 0.369. The validation regimes 
cover a bear, a recovery, and a bull year. 
The S\&P 500 returns $-18.99\%$, 26.00\%, and 25.28\%, respectively.
XGBoost, TabNet, FT-Transformer and the Hybrid return positive in all three, whereas LR is 
negative throughout and MLP negative in 2023.

Only 2025 reflects OOS behaviour. The Hybrid ensemble is the strongest 
performer, returning 51.26\% with a Sharpe ratio of 2.44 and the lowest maximum 
drawdown of any model at $-7.76\%$, exceeding its constituents despite 
trailing XGBoost in 2023 and 2024. XGBoost is the strongest individual model at 
33.61\% and a Sharpe of 1.37. Long and Short precision sits between 
0.223 and 0.254 across all models, with the Hybrid highest on the Short side at 
0.254 and MLP highest on the Long. Confusion matrices in Appendix~\ref{app:conf} show the underlying 
prediction counts. Most observations are assigned to the Hold class alongside above-random 
diagonal counts in both minority classes.

\begin{table}[H]
\centering
\caption{Statistical Robustness and Distribution Analysis -- OOS}
\label{tab:stats_robustness}
\centering
\footnotesize
\renewcommand{\arraystretch}{1.1}
\setlength{\tabcolsep}{6pt}
\begin{tabular}{l cccccc}
\toprule
\textbf{Statistical Metric} & \textbf{LogReg} & \textbf{XGBoost} & \textbf{MLP} & \textbf{TabNet} & \textbf{FT-Trans.} & \textbf{Hybrid} \\
\midrule
\multicolumn{7}{l}{\textit{Risk-Adjusted Outperformance}} \\
\midrule
CAPM $\alpha_\text{ann}$          & -0.071  & 0.276   & 0.208   & 0.107   & 0.037   & \textbf{0.423} \\
CAPM $\beta$                       & 0.112   & 0.200   & -0.154  & -0.031  & 0.047   & 0.048  \\
CAPM $p$-value ($\alpha$ vs S\&P 500)  & 0.603   & 0.171   & 0.264   & 0.479   & 0.833   & \textbf{0.011}$^{*}$ \\
Prob.\ Sharpe Ratio (PSR)         & 0.083   & 0.740   & 0.499   & 0.369   & 0.238   & \textbf{0.960}$^{*}$ \\
\midrule
\multicolumn{7}{l}{\textit{Distribution \& Significance Tests}} \\
\midrule
KS $D$-stat (vs S\&P 500)              & 0.121   & 0.072   & 0.112   & 0.100   & 0.125   & 0.084  \\
KS $p$-value (vs S\&P 500)             & 0.054   & 0.534   & 0.086   & 0.163   & \textbf{0.042}$^{*}$ & 0.339  \\
Wilcoxon $p$-value (vs S\&P 500)       & 0.073   & 0.939   & 0.862   & 0.243   & 0.212   & 0.682  \\
\midrule
\midrule
\textbf{Global Significance} & \multicolumn{6}{c}{Friedman Test $p$-value = 0.352 \hspace{0.2cm} (Stat = 5.555)} \\
\bottomrule
\end{tabular}
\begin{tablenotes}
\footnotesize
\item \textbf{Benchmarks:} CAPM, KS tests, and Wilcoxon signed-rank tests evaluate strategy returns against the S\&P 500 baseline. The PSR utilises the S\&P 500 daily Sharpe ratio as the target.
\end{tablenotes}
\end{table}

Table~\ref{tab:stats_robustness} tests if the Hybrid's outperformance is 
statistically robust or a product of chance. The Hybrid is the only model to 
generate significant CAPM alpha ($\alpha_{\text{ann}} = 0.423$, $p = 0.011$), 
with a near-zero beta ($\beta = 0.048$). Other models' alpha $p$-values 
range from 0.171 (XGBoost) to 0.833 (FT-Transformer), none reaching 
significance. The Hybrid's PSR~\cite{lopezdeprado2018} of 0.960 supports this,
indicating a 96\% probability that its Sharpe ratio 
significantly exceeds the S\&P 500's.

Wilcoxon and KS tests largely fail to reject distributional 
equality against the S\&P 500, the exception being the
FT-Transformer (KS $p = 0.042$), and the Friedman test ($p = 0.352$) does not 
reject equal rank distributions across the six strategies. Pairwise CAPM 
regressions between models are reported in Appendix~\ref{app:stats}.

\begin{figure}[H]
\centering
\includegraphics[width=0.93\textwidth]{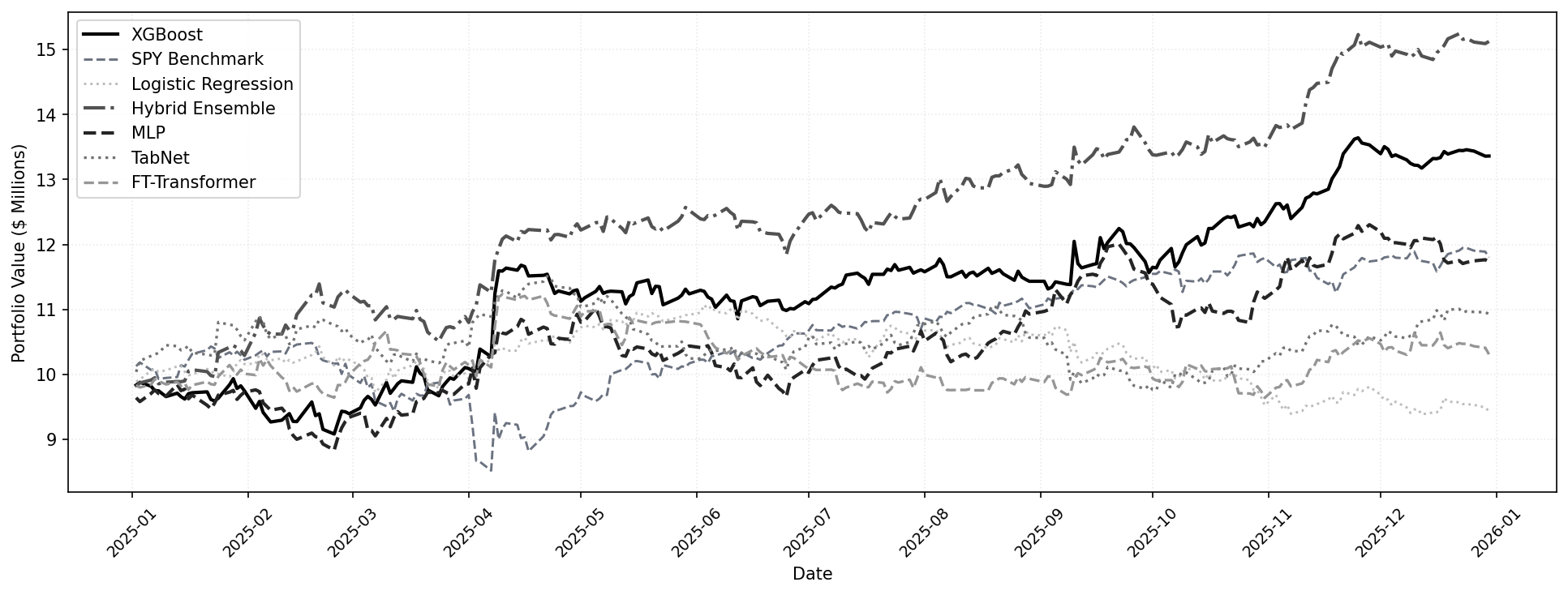}
\caption{Equity Curves of All Models vs S\&P 500 Benchmark -- OOS}
\label{fig:equity_curves}
\end{figure}

Figure~\ref{fig:equity_curves} plots the cumulative portfolio value for all models 
and the S\&P 500 benchmark over the OOS period, starting from a \$10 million 
initial investment. The Hybrid ensemble separates from the field from 
April 2025 onwards, reaching approximately \$15.1 million by year end.
XGBoost is the strongest 
individual model, ending near \$13.4 million, as MLP and TabNet diverge
progressively through the year, finishing at approximately \$11.7 million and 
\$10.9 million respectively. LR and FT-Transformer underperform all other models, 
with LR the only model to end in a loss, finishing at approximately 
\$9.4 million.

A notable feature is the upward step visible across all models in early 
April 2025, coinciding with the S\&P 500's sharp decline following the US tariff 
announcements and its rapid recovery once a pause was 
announced. The Hybrid's low maximum drawdown of $-7.76\%$, reported in 
Table~\ref{tab:trading_performance}, holds despite this period of elevated 
volatility. The S\&P 500 itself performs reasonably 
over the period but is surpassed by XGBoost and the Hybrid in absolute 
terms, and on a risk-adjusted basis only the Hybrid achieves 
statistically significant outperformance as established 
in Table~\ref{tab:stats_robustness}.

\subsection{Robustness and Temporal Stability}

\begin{table}[h]
\centering
\footnotesize
\caption{Hybrid Model Robustness Analysis: Gaussian Noise Perturbation and Null Hypothesis Test}
\label{tab:robustness}
\begin{tabular}{lccccccc}
\toprule
& \multicolumn{4}{c}{\textbf{Gaussian Noise ($\sigma$)}} 
& \multicolumn{2}{c}{\textbf{Reference}} \\
\cmidrule(lr){2-5} \cmidrule(lr){6-7}
\textbf{Metric} & \textbf{0.05} & \textbf{0.10} & \textbf{0.20} & \textbf{0.50} 
& \textbf{Clean} & \textbf{Random Null} \\
\midrule
JS Divergence (Long)  & 0.0126 & 0.0179 & 0.0243 & 0.0350 & -- & -- \\
JS Divergence (Short) & 0.0102 & 0.0129 & 0.0171 & 0.0261 & -- & -- \\
\addlinespace
Total Return (\%) & 25.17 & 34.36 & 2.72  & 6.66  & 51.26  & $-$8.21 \\
Sharpe Ratio      & 1.13  & 1.53  & 0.04  & 0.26  & 2.44   & $-$1.44 \\
Max Drawdown (\%) & $-$7.50 & $-$9.00 & $-$14.23 & $-$9.76 & $-$7.76 & $-$8.47 \\
\bottomrule
\end{tabular}
\begin{minipage}{\textwidth}
\footnotesize\raggedright
\vspace{0.75em}
\textbf{Note}: JS Divergence measures distributional shift between clean and perturbed predicted probabilities; not applicable (--) to Clean or Random Null. Portfolio metrics follow Section~\ref{sec:backtesting}. Random Null replaces all features with Gaussian noise, confirming returns are attributable to learned signal.
\end{minipage}
\end{table}

Table~\ref{tab:robustness} evaluates the sensitivity of the Hybrid model to 
input perturbation across Gaussian noise levels and a random null baseline. 
JS Divergence increases consistently with $\sigma$, rising from 0.0126 (Long) 
and 0.0102 (Short) at $\sigma = 0.05$ to 0.0350 and 0.0261 at $\sigma = 0.50$.
All values remain small in absolute terms, and predicted probability 
distributions are therefore relatively stable under moderate noise. Portfolio 
performance degrades slowly at low noise levels, with $\sigma = 0.10$ still 
producing a Sharpe ratio of 1.53 and a total return of 34.36\%, before 
decreasing sharply at $\sigma = 0.20$. The 
Random Null baseline yields a total return of $-8.21\%$ and a Sharpe ratio 
of $-1.44$. This rules out backtest mechanics and random chance as 
explanations for the Hybrid's OOS performance. Full degradation and 
divergence curves are plotted in Appendix~\ref{app:robustness_figures}.

\begin{table}[h]
\centering
\footnotesize
\caption{Quarterly Temporal Stability Analysis: Signal Precision Across 2025 Out-of-Sample Sub-Periods}
\label{tab:temporal_stability}
\begin{tabular}{lcccccc}
\toprule
& \multicolumn{4}{c}{\textbf{Signal Precision (\%)}} 
& \multicolumn{2}{c}{\textbf{Observations}} \\
\cmidrule(lr){2-5} \cmidrule(lr){6-7}
\textbf{Quarter} & \textbf{Long} & \textbf{Short} & \textbf{Long vs Baseline} & \textbf{Short vs Baseline} 
& \textbf{N Long} & \textbf{N Short} \\
\midrule
Q1 (Jan--Mar) & 23.06 & 26.67 & $+$13.06 & $+$16.67 & 360 & 360 \\
Q2 (Apr--Jun) & 23.12 & 25.81 & $+$13.12 & $+$15.81 & 372 & 372 \\
Q3 (Jul--Sep) & 25.78 & 25.00 & $+$15.78 & $+$15.00 & 384 & 384 \\
Q4 (Oct--Dec) & 24.07 & 24.34 & $+$14.07 & $+$14.34 & 378 & 378 \\
\bottomrule
\end{tabular}
\begin{minipage}{\textwidth}
\footnotesize\raggedright
\vspace{0.75em}
\textbf{Note}: Precision is the proportion of predicted Long/Short signals correctly identifying the true class. Random baseline is 10\% (outer-decile construction). Long vs Baseline and Short vs Baseline show percentage-point improvement over the 10\% random baseline. Results reported for the Hybrid model.
\end{minipage}
\end{table}

Table~\ref{tab:temporal_stability} examines whether the Hybrid model's predictive 
signal is stable across the four quarters of the OOS period. Long 
and Short precision exceed the 10\% random baseline by approximately 
13--17 percentage points in every quarter, and no quarter shows a significant 
collapse in performance. The model does not rely on a single favourable 
market regime to generate its signal as long precision ranges narrowly from 23.06\% (Q1) to 
25.78\% (Q3), and Short precision from 24.34\% (Q4) to 26.67\% (Q1). 
As KS tests confirm the quarterly return and volatility distributions 
are statistically distinct ($p < 0.001$ for all pairs), it suggests 
that the model may be capturing more generalisable patterns that are robust to changing 
market conditions, without overfitting to specific temporal dynamics.
Quarterly precision is plotted against the random baseline in 
Appendix~\ref{app:robustness_figures}.

\subsection{Feature Importance and SHAP Analysis}\label{sec:shap_results}

\begin{figure}[H]
    \centering
    \includegraphics[width=0.8\textwidth]{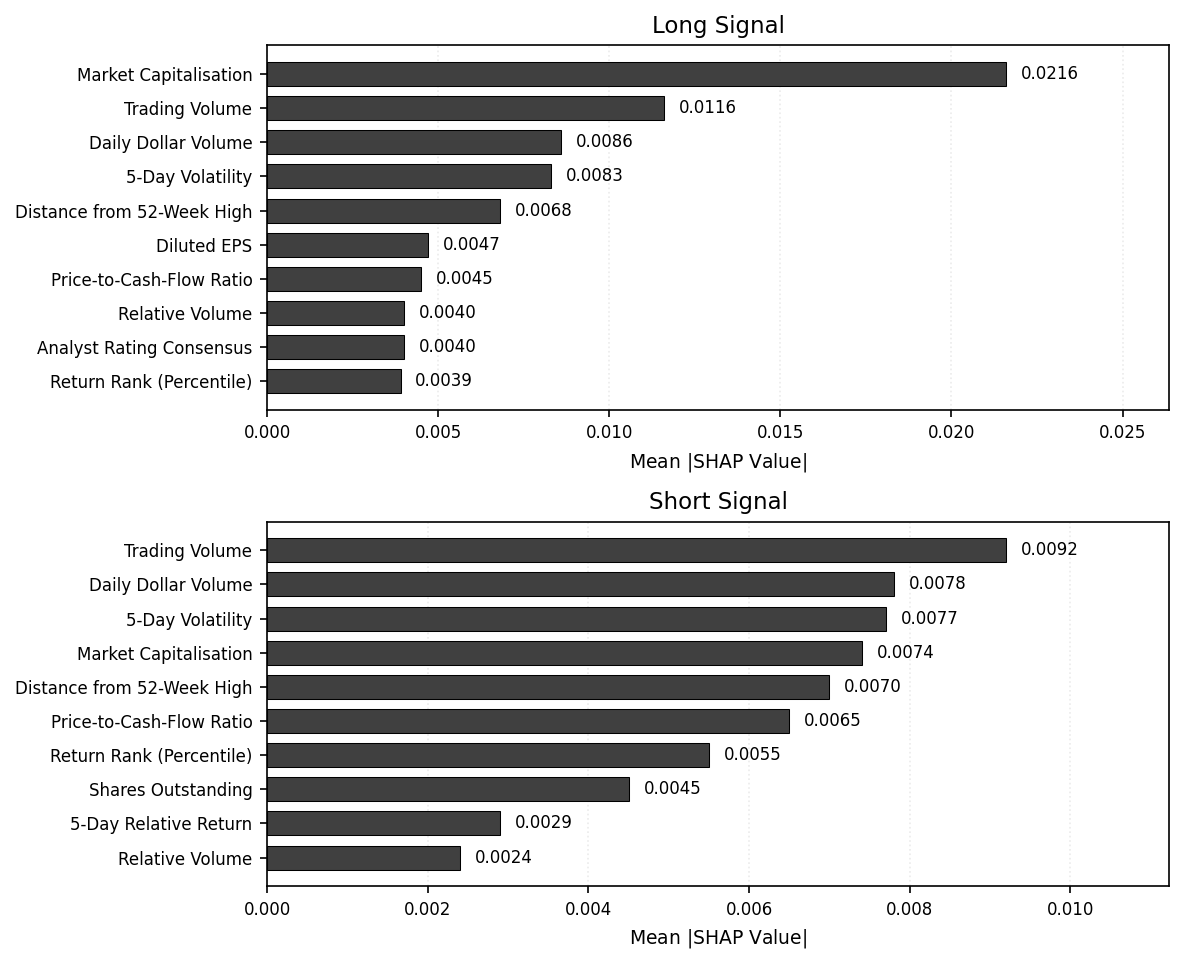}
    \caption{Top 10 features by mean SHAP value for the long and short signals -- XGBoost}
    \label{fig:xgb}
\end{figure}

\begin{table}[h]
\footnotesize
\centering
\caption{Total Mean SHAP Contribution by Feature Category Across All Models}\label{tab:shap_category_all}
\begin{tabular*}{\textwidth}{@{\extracolsep{\fill}}llcccc@{}}
\toprule
Model & Signal & Technical & Fundamental & Macroeconomic & Alternative \\
\midrule
XGBoost & Long  & 50.87\% & 43.73\% & 3.76\%  & 1.63\% \\
        & Short & 59.45\% & 33.30\% & 4.36\%  & 2.89\% \\
\addlinespace
LR & Long  & 55.93\% & 14.44\% & 15.56\% & 14.08\% \\
   & Short & 63.09\% & 11.89\% & 12.82\% & 12.20\% \\
\addlinespace
MLP & Long  & 40.94\% & 32.98\% & 13.52\% & 12.56\% \\
    & Short & 47.51\% & 27.17\% & 15.16\% & 10.16\% \\
\addlinespace
TabNet & Long  & 58.97\% & 18.41\% & 15.43\% & 7.19\% \\
       & Short & 57.79\% & 21.60\% & 12.12\% & 8.48\% \\
\addlinespace
FT-Transformer & Long  & 52.58\% & 30.30\% & 8.90\%  & 8.22\% \\
               & Short & 55.94\% & 27.01\% & 9.43\%  & 7.62\% \\
\bottomrule
\end{tabular*}
\end{table}

Figure~\ref{fig:xgb} reveals that XGBoost relies predominantly on technical 
and market-based features. Market Capitalisation leads the Long signal at 
0.0216, nearly double Trading Volume's 0.0116, but falls to fourth on the 
Short signal (0.0074) as Trading Volume takes first place, confirming the 
model has learned asymmetric decision boundaries rather than a simple signal 
reversal. 

At the category level, technical features lead both directions, 
exceeding fundamentals by 7 percentage points on the Long signal 
and 26 points on the Short. Macroeconomic and alternative features together 
account for under 6\% of attribution in both directions, indicating a 
supplementary role. Per-architecture SHAP breakdowns for all five 
base models are provided in Appendix~\ref{app:shap}.

\subsection{Research Application and Portfolio Sensitivity}

\begin{figure}[H]
\centering
\includegraphics[width=0.9\textwidth]{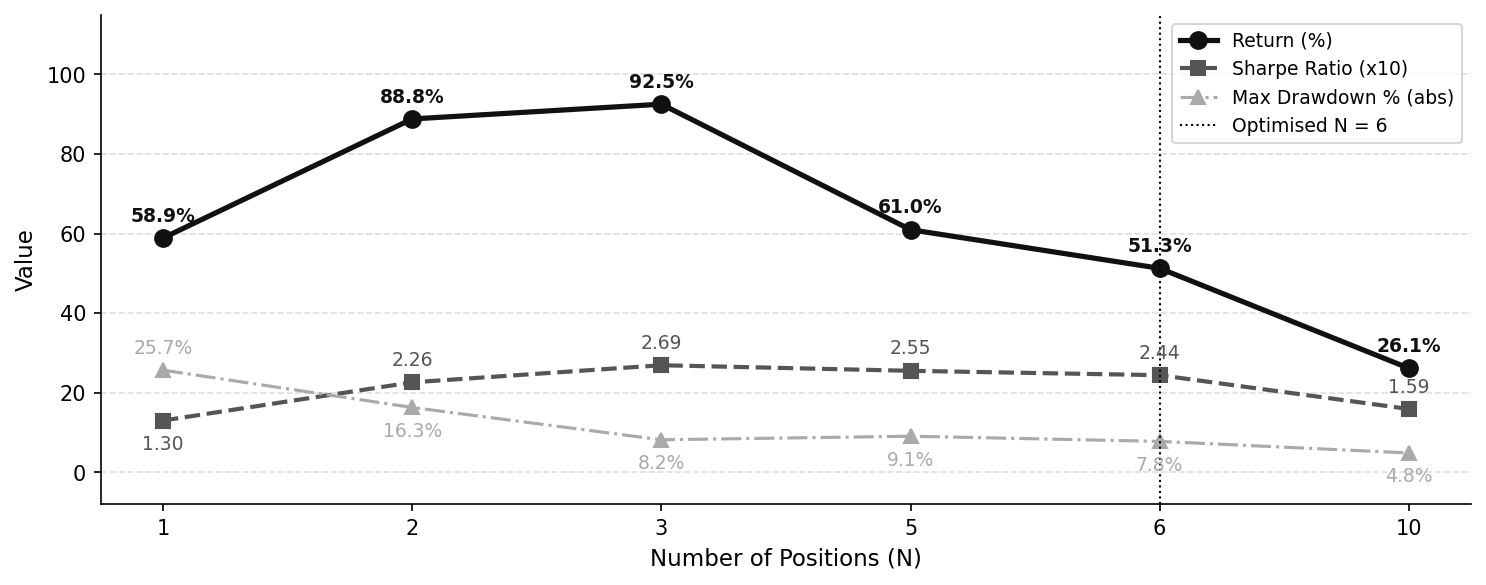}
\caption{Hybrid Ensemble Performance Sensitivity to Position Size -- OOS}
\label{fig:n-sensitivity}
\end{figure}

Figure~\ref{fig:n-sensitivity} illustrates how portfolio performance varies 
with the number of stock positions~$N$. Returns and Sharpe ratio peak 
at $N = 3$, falling to 26.1\% by $N = 10$ as signal 
dilution from lower-ranked positions outweighs diversification benefits. 
Maximum drawdown moves inversely, improving from $-$25.7\% at $N = 1$ to 
$-$4.8\% at $N = 10$ as concentration risk falls. The Bayesian-optimised 
configuration of $N = 6$, Sharpe 2.44, drawdown $-$7.76\%, trades some peak 
return for stability across a broader set of positions. This sensitivity 
analysis is directly replicable within the interactive research application,
shown in Appendix~\ref{app:application}.

\section{Discussion}
\subsection{Model Comparison and Interpretation}

Aggregate classification metrics are a poor indicator of model 
quality, as shown in Table~\ref{tab:trading_performance}. 
The structural class imbalance introduced by the 
outer-decile target construction is important to understanding why. 
80\% of observations are assigned to the Hold class, therefore accuracy and F1 
are dominated by majority-class performance. A model predicting hold 
for every observation would achieve 80\% accuracy 
without generating a trading signal. The narrow 
accuracy range of 0.786 to 0.799 and F1 scores of 0.337 to 0.369 
observed across all models demonstrates this pattern. The practical 
consequence is most visible in the difference between LR and XGBoost, 
which report near-identical aggregate accuracy of 0.789 and 0.794, 
respectively, but produce trading returns of $-5.69\%$ and $33.61\%$ 
in the OOS period. This disconnect arises because the hold 
class contributes nothing to portfolio returns by construction. Only 
the most confident long and short predictions are used to assign trading positions. A model's 
aggregate accuracy therefore reflects its ability to 
identify stocks that will not be traded. 

Even at the per-class level, LR achieves 
short precision of $0.240$ and long precision of $0.233$, both 
comparable to XGBoost's $0.239$ in each direction, indicating that 
even per-class precision cannot capture the signal quality differences 
that determine portfolio performance.
This is potential evidence of the mechanism raised in Section~\ref{sec:regime_eval}, where
precision weights every correct or incorrect position equally regardless 
of the magnitude of the return, whereas portfolio performance 
is magnitude-weighted. Two models can therefore share near-identical precision 
but one's correct trades concentrate on the largest-magnitude moves and 
the other's on marginal ones. Over time this can materialise as 
large differences in realised portfolio returns.
This finding reinforces Lopez De Prado~\cite{lopezdeprado2018} and Fieberg et al's~\cite{Fieberg2023} 
arguments that models ranking highest on predictive accuracy do not 
consistently produce the strongest long-short portfolio returns. 

Within the individual models, XGBoost is the strongest single model 
in the OOS period with a return of $33.61\%$ and a Sharpe ratio 
of 1.37. The dataset contains approximately 750,000 stock-day observations 
across only around 2,500 unique trading days, so the effective number of 
independent market environments available for learning is far smaller than 
the observation count suggests. The gradient-based optimisation 
embedded within deep learning models typically needs a larger effective 
sample to converge than tree-based methods, along with more tuning 
and computation to reach comparable 
performance~\cite{Grinsztajn2022, Borisov2024, ShwartzZiv2022}. These efficiency 
differences could plausibly explain part of XGBoost's edge over the three 
deep learning architectures.
The FT-Transformer illustrates this cost most clearly, 
achieving the weakest deep learning return at 3.08\% despite 
its greater complexity. Its 
inconsistent cross-validation returns across the 
three regimes imply its attention mechanism 
likely identifies regime-specific correlation, the exact failure mode that cross-regime optimisation was 
designed to prevent. TabNet performs better than the FT-Transformer, 
but still falls short of XGBoost's performance, indicating that architectural 
similarity in principle does not produce equivalent performance in deployment. 
TabNet's sparsemax layers were designed to mimic tree-based 
node splitting, but this has not closed 
the performance gap with XGBoost's boundary splits in this environment.

Each TDL model was allocated the same 30-trial budget as the baselines. Deep 
learning models are more sensitive to hyperparameter settings 
than gradient-boosted trees, so this budget
may have been insufficient to identify a configuration that generalises across regimes.
Grinsztajn et al.~\cite{Grinsztajn2022} examine this by 
evaluating performance across search budgets and find 
tree-based models are superior in each one, suggesting 
the ordering observed here is unlikely to reverse under a larger 
search. However, their benchmark was not related to this research so direct comparison 
is an avenue for future work.

The Hybrid ensemble is the only model to generate statistically 
significant OOS alpha, despite substantial differences in 
performance across the individual models. This advantage comes from 
combining models whose errors are uncorrelated. Each model's mistakes 
are likely to be offset by the other's correct prediction on the same 
stock-day, making the combined signal more often 
correct~\cite{Dietterich2000, Ortega2021}. The composite scoring 
function, described in Section~\ref{sec:ensemble}, selected TabNet as 
XGBoost's ensemble partner on this basis, despite MLP's stronger 
performance. The pairwise CAPM $R^2$ is $0.028$ between 
XGBoost and TabNet, far lower than the $0.149$ between XGBoost and 
MLP, shows that TabNet's errors are less correlated with XGBoost's 
than MLP's are. Preserving this orthogonality in the combined signal 
depends on how the two models' are aggregated. Averaging probabilities risks an overconfident but 
poorly calibrated model dominating the score. On the other hand, ranking 
each model's output before averaging expresses every prediction on the 
same ordinal scale. That rank aggregation producing significant alpha 
without a trained meta-learner suggests simple combination can capture 
signal orthogonality. Regime-conditional weighting could improve on this in future research.

Statistical testing indicates the Hybrid's outperformance is not due to chance. 
The CAPM alpha and near-zero beta indicate stock selection skill, separate from 
market exposure. The PSR's $96\%$ probability of genuine outperformance supports this, 
once skewness and kurtosis are accounted for. In contrast, the KS and Friedman tests do not 
detect this, finding the Hybrid's daily return distribution not significantly different in
shape from the S\&P 500's and no difference in daily rank ordering 
across strategies. These tests assess the typical day, so a strategy can 
pass both but still find its alpha from a small number of large-magnitude days. 
An example of this is the upward step visible in Figure~\ref{fig:equity_curves} 
during the April 2025 US tariff volatility. The S\&P 500 fell on the 
announcement, but the Hybrid's near-zero beta meant it was not exposed to 
the market-wide decrease. Under dollar-neutral construction, a gain on the short book can cause 
a net portfolio increase. The Hybrid's short precision of $0.254$, sustained 
across the full OOS period, suggests this particular gain reflected short-side 
signal quality.  More broadly, this short precision is the highest of any model 
and corresponds with the lowest maximum drawdown across all models at $-7.76\%$, 
implying a relationship between short-side signal quality and downside 
protection, though the analysis here cannot isolate the precise contribution. 

The robustness analysis shows the Hybrid tolerates realistic input 
degradation. Performance collapses at $\sigma = 0.20$, a 
level likely more severe than data quality issues that would 
typically stem from vendor inconsistencies or pipeline latency in 
production; this is discussed further in Section~\ref{sec:deployment}. The null baseline result is the more important finding here. 
When all input features are replaced with noise, the Hybrid's Sharpe 
ratio falls to $-1.44$, compared to $2.44$ on real data, ruling out 
backtest mechanics or random chance as an explanation for its OOS 
performance and leaving learned signal as the remaining explanation. 
Signal precision remains stable within a narrow range across all 
four distinct quarters of the OOS period, well above the 
$10\%$ random baseline, with no quarter showing a significant collapse. 
Conversely, precision is an imperfect indicator of trading value, so the more 
meaningful observation is that this stability holds across conditions 
the model was not trained on. Input robustness and temporal 
stability together imply the Hybrid captures generalisable market 
structure, though future work evaluating longer OOS horizons could 
strengthen this conclusion.

TDL architectures do not provide individual 
improvements over gradient-boosted trees in this setting, reiterating 
the conclusions of Grinsztajn et al~\cite{Grinsztajn2022} and Borisov et al~\cite{Borisov2024} in an 
equity context. TabNet's contribution appears through signal orthogonality with XGBoost.
The Hybrid's statistically significant 
alpha and sustained quarterly precision provide evidence that cross-regime Bayesian 
optimisation is likely producing hyperparameters 
that generalise beyond the conditions under which they were 
estimated; robustness to realistic levels of input 
perturbation offers further support. These results also outperform 
Pagliaro~\cite{Pagliaro2026}, who reports a Sharpe of $1.18$ using a 
regime-aware LightGBM framework. The performance gap suggests that an 
ensemble combination in the cross-regime framework may produce a more 
generalisable signal than a single regime-aware model. Further 
examination is required as the two studies use different datasets and 
evaluation periods. The single OOS year evaluated here limits 
the strength of any such comparison, a point returned to in 
Section~\ref{sec:limitations}.

\subsection{Value of Alternative Data}\label{sec:alt_data}

SHAP analysis, presented in Figure~\ref{fig:xgb} and Appendix~\ref{app:shap}, 
reflects the category frequency found by Kumbure et al.~\cite{Kumbure2022}. 
Technical indicators dominate across all models at 40--63\%, 
reiterating Gu et al.~\cite{gu2020} who find momentum and price-based 
features among the most important predictors. 
The split between remaining categories varies considerably by architecture. 
XGBoost shows $43.73\%$ attribution to fundamental features but only 
$1.63\%$ to alternative data, with LR showing a different balance at 
$14.44\%$ and $14.08\%$, respectively. LR's inability to extract 
non-linear interactions from technical and fundamental features 
means sentiment and macroeconomic signals receive attribution that 
those interactions would if present. Conversely, XGBoost captures the 
same information through interaction splits, 
leaving alternative data with little marginal contribution. 
The Unemployment Trend, a search-based 
uncertainty proxy, appearing in LR's top 
features but not XGBoost's illustrates this. LR relies on the 
linear component of search volume signals as a proxy for market 
conditions, a signal XGBoost handles through non-linear splits on 
technical and fundamental variables. This pattern suggests 
attribution results are not comparable across model classes, and 
conclusions about alternative data value derived from one architecture 
may not hold in another. 
Existing studies on alternative data value predominantly use 
linear or regression-based frameworks, as evidenced by the 
studies reviewed in Section~\ref{sec:data_sources}~\cite{Allen2019, 
Gambarelli2025}, meaning the architecture-capacity effects 
identified here may not have been previously observable. The attribution figures 
reported here are themselves architecturally dependent and should 
be treated as a representation of signal importance within 
each model class. KernelExplainer with 5,000 samples and 100 k-means 
background centres approximates Shapley values, and low-attribution features in alternative 
and macroeconomic categories should be interpreted with this 
in mind. To discover whether the low attribution reflects genuine signal value 
or architectural capacity constraints, ablation studies testing category 
combinations across model classes are needed. Until then, alternative 
data attribution should be evaluated within the specific model class 
intended for deployment.

Low alternative data attribution in the strongest performing models does not 
imply these sources lack predictive value. XGBoost, the strongest individual 
model, falls well short of the 10\% threshold, while LR and 
MLP comfortably exceed it, shown in Table~\ref{tab:shap_category_all}. The hypothesis is 
therefore rejected for the best-performing architecture but not uniformly 
across model classes, reinforcing that alternative data's apparent contribution 
is architecture-dependent. Sentiment and search attribution may also be understated 
because news typically moves prices on the day of release, so the one-day lag applied 
in preprocessing, discussed in Section~\ref{sec:preprocessing}, risks discarding same-day signal~\cite{Briere2023}. 
This effect could be analysed within further studies by moving to an intraday or high-frequency trading 
setting where sentiment and price data can be aligned at finer granularity. 
Even accounting for this, sentiment and search-based features capture information 
orthogonal to price dynamics and accounting fundamentals~\cite{SunYunchuan2024, Gambarelli2025, Atkins2018}, 
meaning their contribution is additive even when small in absolute terms. This supplementary 
role is consistent with evidence that sentiment is priced in equity returns without 
displacing technical or fundamental signals~\cite{Gambarelli2025}, and suggests that 
alternative data sources should be evaluated for their diversifying properties. 
A marginal attribution in this range could influence which stocks sit 
at the boundaries of the top and bottom deciles, but quantifying this 
effect would require targeted ablation studies removing alternative features entirely.

Alternative data generally contributes more to short-side signals than 
long-side signals, as shown in Table~\ref{tab:shap_category_all}. 
This is consistent with Tetlock~\cite{Tetlock2007}, who shows that 
media pessimism predicts downward price pressure. Negative signals 
are more relevant to returns than equivalent positive ones, 
as investors discount good earnings news but react sharply to bad news 
when uncertainty is elevated~\cite{Bird2012}, and rising search 
volumes are a marker of exactly this kind of 
uncertainty~\cite{Szczygielski2024}. Stocks experiencing negative news 
therefore generate more identifiable signal from sentiment and 
search-based features than stocks experiencing positive news of 
equivalent magnitude, matching the greater alternative-data 
contribution to short-side relative to long-side predictions. 
Technical features show the same directional pattern, with short 
signal attribution consistently higher than long signal attribution. 
Hong and Stein~\cite{HongStein1999} model gradual information 
diffusion producing short-run underreaction and momentum, an effect 
Hong et al.~\cite{Hong1998} find operates more strongly for past 
losers than past winners. If bad news diffuses more slowly through 
the market than good news, price trends among declining stocks should 
be more persistent and therefore more predictable than trends among 
rising stocks, explaining why technical features, built from recent 
price and momentum data, carry more weight in the models' short-side 
predictions than their long-side predictions. However, this is contentious, 
as Gambarelli and Muzzioli find that negative news is incorporated into prices more 
quickly than positive news in European equities~\cite{Gambarelli2025}, suggesting that the asymmetry 
observed here may be specific to US equities in this OOS period. 
Future research examining if this asymmetry exists within 
different market regimes or alternative universes would determine how 
robust this finding is beyond the OOS period.

\subsection{Practical Deployment Considerations}\label{sec:deployment}

The sensitivity analysis in Figure~\ref{fig:n-sensitivity} shows 
peak Sharpe at $N=3$. In contrast, the Bayesian-optimised 
configuration selects $N=6$. Concentrating capital into fewer 
positions amplifies high-conviction predictions when signal quality 
is high, but as $N$ increases, stocks ranked lower in the model's 
rankings are added to the portfolio and signal dilution outweighs 
any benefits from diversification. However, this peak at $N=3$ is not practically 
accessible. Concentration at this level would breach conventional 
issuer concentration limits in traditional asset management 
mandates~\cite{Chen2025}. At institutional scale, executing 
a three-stock position would generate market impact sufficient 
to reduce the targeted returns~\cite{lopezdeprado2018, Zovko2020}. 
Even at $N=6$, the strategy exceeds 
concentration thresholds, making it more suitable for 
proprietary trading desks or hedge funds with flexible mandates 
than benchmark-constrained investors.

The robustness analysis identifies a performance threshold 
at $\sigma=0.20$, after which risk-adjusted performance collapses to a 
level not worth the operational cost of deployment, illustrated in 
Table~\ref{tab:robustness}. This marks the point 
where input quality impairs signal generation. However,  
the Gaussian noise injection likely understates the true 
risk. Real-world failures rarely 
appear as random noise. A vendor reformatting its data feed or
returning null and incorrect values would each corrupt specific 
inputs while others would be unaffected. Structured 
missingness may be harder to detect and more 
damaging than the Gaussian perturbations simulated, since 
errors in specific inputs can silently distort the model's 
predictions without triggering obvious warning signals. 
The distributional shift between the training period 
and OOS, shown in Appendix~\ref{app:ks}, raises a related concern. 
The model was already operating on a shifted 
distribution when it reached the OOS period, 
and the quarterly distributional analysis confirms that regime 
transitions occur within the span of a single deployment year. 
Annual retraining is therefore likely insufficient under live 
conditions, particularly if market structure shifts mid-year 
in ways a fixed schedule cannot anticipate. A production system 
would require continuous monitoring for distributional drift, 
using statistical tests similar to those applied here for 
regime validation, triggering retraining when significant shifts 
are detected~\cite{Klaise2020, Pagliaro2026}. Tuning 
and execution time also determine which retraining cadence is 
operationally viable. XGBoost's full 30-trial tuning run completed in 
under 11 minutes, comfortably supporting daily retraining within a 
single overnight close, whereas TabNet's took approximately 19.6 hours, 
fitting within the weekend market closure but ruling out daily 
retraining at this trial count. Full tuning, training, and inference 
times for all models are reported in Appendix~\ref{app:runtime}. 
Inference times reported there cover the full 252-day OOS 
period. Per trading day, this corresponds to under 60 milliseconds even 
for TabNet, the slowest model, which is well within the time available before 
market-on-close execution. This headroom suggests the modelling 
pipeline is fast enough to support higher-frequency signal generation, 
but this depends on data feed and feature computation latency, 
neither of which was evaluated here.

Most research presents findings as fixed outputs, 
preventing practitioners from testing assumptions or examining 
model behaviour. The interactive 
application addresses part of this. Position sizing sensitivity 
is explorable in real time. However, the simulated configurations 
do not account for the market impact that makes concentrated 
positions impractical at institutional scale, and transaction 
cost sensitivity remains fixed at 2.2 basis points.
The SHAP attribution interface exposes why 
signals are generated across models and 
directions~\cite{Lundberg2017}, supporting governance 
requirements in professional deployment~\cite{Finra2026}.
Backtesting results and SHAP attribution are not straightforward for 
non-specialist users to understand. The AI research 
assistant addresses this by allowing natural language questions over 
both, without requiring technical expertise or pipeline access.
The application runs on historical data and 
attribution derived from a fixed training distribution may 
not hold under the regime shifts confirmed between training 
and OOS. Widespread adoption of similar strategies could 
add to the market-impact effects discussed across 
institutions, potentially amplifying 
volatility during regime transitions, a risk relevant to any 
production deployment of this framework. 

\subsection{Limitations}\label{sec:limitations}

Using S\&P 500 constituents identified at the end of the sample period introduces 
survivorship bias. Stocks delisted or removed before 2025 are excluded,
and as these tend to be underperformers, backtested returns are overstated relative 
to a point-in-time universe~\cite{Brown1992}. All six strategies trade the 
same universe, so the bias affects the absolute return level but the relative 
ordering across models is unaffected. However, its precise magnitude 
cannot be determined without reconstructing the universe using historical 
index membership data, providing an avenue for future research. Another 
limitation is that data was sourced through a 
Bloomberg Terminal, which requires an institutional licence, limiting 
independent replication of these results without equivalent access. 
Further studies could substitute publicly available sentiment and fundamental 
sources to test whether performance depends materially on this licensed data.

The OOS evaluation covers 252 trading days in 2025, limiting the 
statistical power available to determine if performance 
differences across models are significant. The Friedman test's 
failure to reject equality across strategies partly demonstrates this, 
and a formal power calculation would clarify the minimum detectable 
effect size at this sample size. Also, reported figures reflect the 
selection of the best-performing configuration from approximately 
161 trialled across the five model classes and their ensemble 
combinations. Deflated Sharpe Ratio~\cite{lopezdeprado2018} would correct
for this, but was not applied here as per-trial 
Sharpe ratios were not retained during optimisation. The PSR reported in 
Table~\ref{tab:stats_robustness} should therefore be read as an 
upper bound on risk-adjusted significance. 
The scale of this effect is reduced because selection and 
reporting used different data, as configurations were chosen on the 
validation folds and all reported figures derived from 
the held-out period. Published 
predictive signals also tend to decay as market participants learn 
to exploit them~\cite{Mclean2016}, and the OOS results show a 
period before widespread adoption of this specific approach, meaning 
future performance cannot be assumed to remain at the same level. 
Rolling OOS evaluation across multiple years would provide a more 
robust test of if the strategy's performance generalises beyond
a single market environment.

No single-regime baseline was tuned under an identical objective 
function. This study therefore evaluates if configurations 
selected under a cross-regime objective generalise to unseen market 
conditions, rather than if they outperform conventional 
single-regime tuning. Establishing the latter requires the same 
search run under both conditions with trial count, search space, 
and evaluation protocol unchanged, and is a direct extension of 
this work.

Short selling introduces costs not captured in the 2.2 basis 
point assumption. Borrowing incurs 
fees that vary across stocks and over time, and availability 
constraints can prevent short positions from being established 
or maintained~\cite{lopezdeprado2018}. These costs tend to be 
highest in high-volatility regimes, exactly when short 
signals are likely to be strongest, meaning the strategy's 
short-side returns are overstated relative to what would be 
achievable in practice. Modelling stock-specific borrow 
costs within the backtesting framework would give a more 
realistic estimate of net short-side performance.

As stated in Section~\ref{sec:alt_data}, SHAP values were approximated due to  
computational constraints that introduce 
approximation error into the attribution estimates. Features 
with low mean absolute SHAP values, particularly in the 
alternative and macroeconomic categories, are most sensitive 
to this approximation and should be interpreted with caution. 
The background dataset was derived from the full
training distribution, meaning attribution estimates reflect 
average feature importance across all regimes without a view on 
regime-specific patterns. Exact Shapley value computation 
would be more reliable but is computationally prohibitive at 
this scale, and future studies using TreeExplainer for tree-based 
models alongside KernelExplainer for deep learning models 
would reduce approximation error and maintain
cross-architecture comparability.

\section{Conclusion}

Cross-regime Bayesian optimisation of a TDL ensemble provides statistically significant 
OOS alpha that no constituent model achieves, demonstrating that 
the value of these architectures in cross-sectional trading strategy is combinatorial. 
Beyond this performance result, the optimisation framework itself 
constitutes a methodological contribution, since treating regime heterogeneity as a constraint 
within hyperparameter selection offers a transferable 
protocol for model tuning under distribution shift that is independent of the specific architectures 
or asset class. The Hybrid ensemble's returns are driven almost 
entirely by stock selection, and this risk-adjusted 
performance exceeds Pagliaro's regime-aware LightGBM framework~\cite{Pagliaro2026}. This suggests cross-regime 
ensemble combination captures a more generalisable signal than a single regime-aware model, 
though isolating this framework's specific contribution from conventional tuning remains a direction for future work. 
Institutions evaluating tabular deep learning for trading should prioritise ensemble 
design and regime-aware tuning protocols.

TabNet and FT-Transformer do not produce significant improvements over gradient-boosted trees, 
aligning with Grinsztajn et al.~\cite{Grinsztajn2022} and Borisov et al.~\cite{Borisov2024} as 
XGBoost remains the strongest individual model. 
TabNet is selected as the ensemble partner over the stronger-performing MLP due to 
signal orthogonality, extending Fieberg et al's.~\cite{Fieberg2023} finding that no single model dominates.
The Hybrid ensemble's OOS performance is robust to realistic levels of input perturbation, 
as performance remains greater than the S\&P 500. Signal precision also remains 
consistently above the random baseline across all four quarterly regimes, indicating 
the predictive edge does not decay within the test year.

Alternative data functions as a supplementary signal in this prediction setting despite the breadth 
of existing research suggesting otherwise. The strongest individual model relies on alternative 
features only marginally, falling short of the hypothesised threshold. Attribution varies considerably 
across architectures, with linear models leaning on alternative data far more heavily than 
tree-based methods. This indicates that alternative data's usefulness depends more 
on how a model represents feature interactions than on the informational content of the 
data itself. Consistent with Bird and Yeung~\cite{Bird2012} 
and Hong and Stein~\cite{HongStein1999}, alternative data contributes more to short-side signals than 
long-side signals, reflecting behavioural asymmetries in how negative information is priced.

Finally, a strategy can be statistically indistinguishable from its peers and still be the most profitable 
in production, since classification metrics being consistent across models does not mean the strategies 
are equivalent in practice. Model selection should therefore be driven by portfolio-level backtesting, 
and ensemble construction should prioritise signal orthogonality over individual model performance.

\newpage

\section*{Acknowledgements}

I would like to thank my supervisor, Dr Gaojie (Jay) Jin, for his guidance
and feedback throughout this project. His input on the experimental design
and the framing of the evaluation was invaluable, and this work is
considerably stronger for it. I am also grateful to the Department of
Computer Science at the University of Exeter for providing access to the
computational resources and licensed data on which this research depends.

\begin{spacing}{0.9}
    \bibliographystyle{IEEEtran}
    \bibliography{report_ref}   
\end{spacing}

\newpage

\appendix

\section{UMAP Feature Space Projections}\label{app:umap}

The following figures present UMAP projections of the training feature space, 
included as visual evidence for the methodological choices discussed 
in Section~\ref{sec:method}. Both figures display the same sample of 
approximately 6,000 observations, stratified by class and year across 2015--2024, 
with each subplot coloured by raw next-day return magnitude (left) and 
cross-sectional decile label (right).

\begin{figure}[H]
    \centering
    \includegraphics[width=\textwidth]{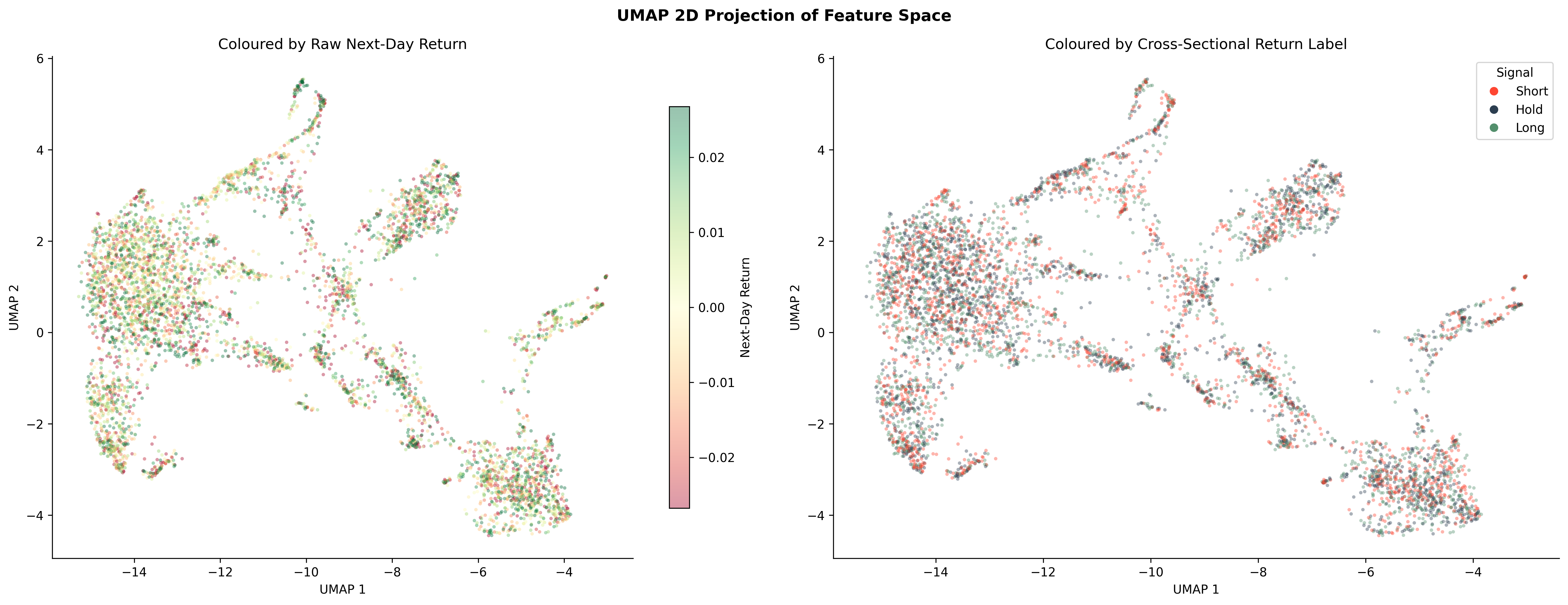}
    \caption{UMAP 2D projection of the feature space coloured by raw next-day 
    return (left) and daily cross-sectional return decile label (right).}\label{fig:umap_2d}
\end{figure}

\begin{figure}[H]
    \centering
    \includegraphics[width=\textwidth]{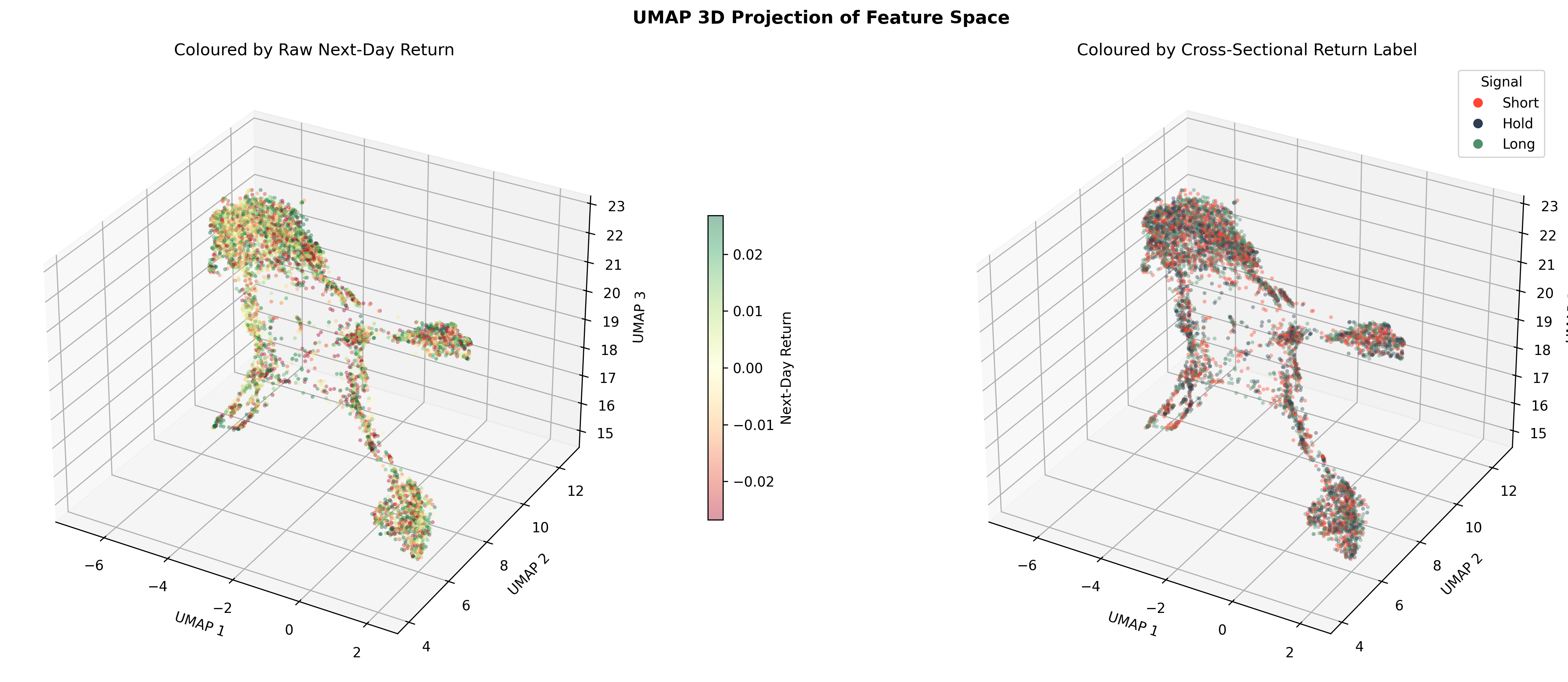}
    \caption{UMAP 3D projection of the feature space coloured by raw next-day 
    return (left) and daily cross-sectional return decile label (right).}\label{fig:umap_3d}
\end{figure}

The absence of systematic colour organisation in both projections signal that absolute 
return magnitude and daily cross-sectional decile labels are not learnable from individual 
feature vectors in isolation, motivating the use of a cross-sectional ranking approach at the portfolio 
construction stage. The three structural clusters visible in the projection may correspond to bear, 
recovery, and bull market regimes, providing unsupervised geometric 
visualisation of the market regime structure identified statistically by 
KS testing in Section~\ref{sec:method}, and justifying the 
multi-regime validation framework.

\newpage 

\section{Kolmogorov-Smirnov Tests and Distributional Analysis}\label{app:ks}

\begin{figure}[H]
    \centering
    \includegraphics[width=\textwidth]{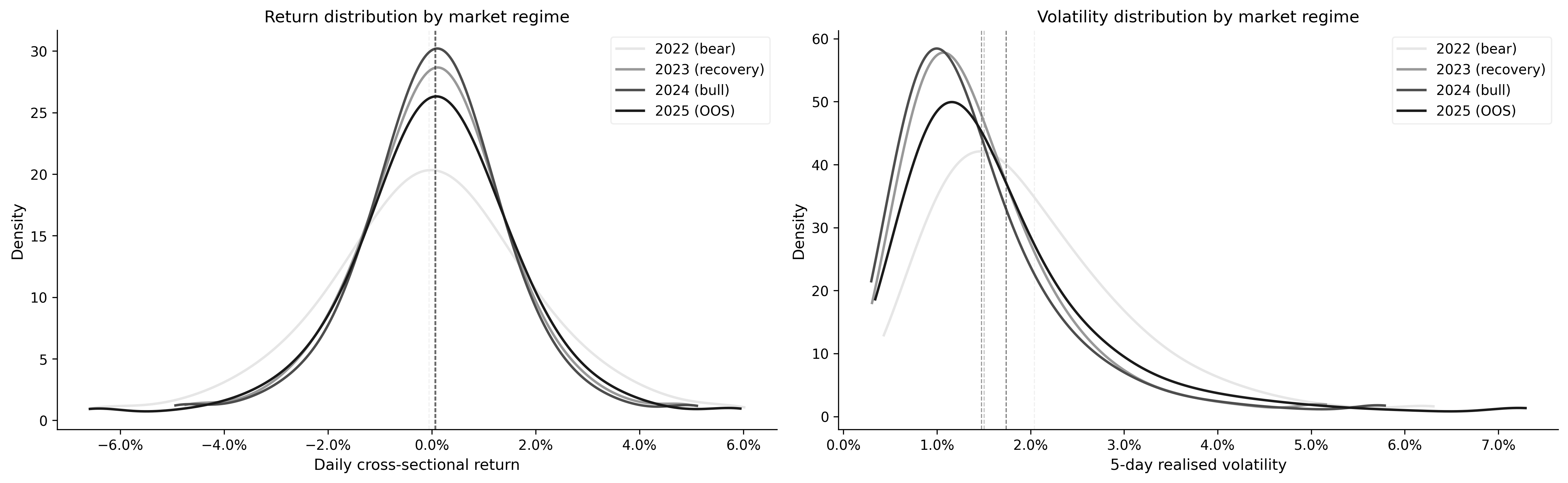}
    \caption{\textbf{Kernel Density Estimates of Return and Volatility Distributions Across Market Regimes (2022--2025).}}\label{fig:regime_kde}
\end{figure}

\begin{figure}[H]
    \centering
    \includegraphics[width=\textwidth]{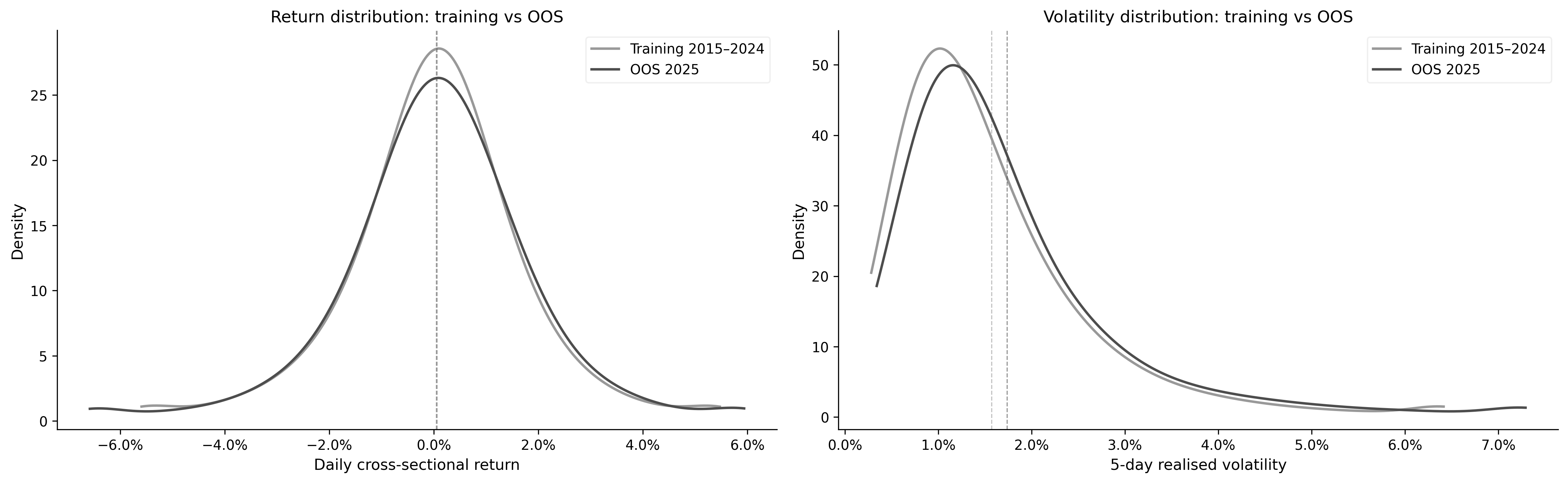}
    \caption{\textbf{Kernel Density Estimates Comparing Training (2015--2024) Against the OOS Period (2025).}}\label{fig:train_oos_kde}
\end{figure}

To assess whether the validation folds correspond to distinct market environments, pairwise two-sample 
KS tests were conducted on daily cross-sectional return distributions and 5-day realised volatility 
distributions. The results are summarised in Table~\ref{tab:ks_tests}, while Figures~\ref{fig:regime_kde} 
and~\ref{fig:train_oos_kde} provide kernel density comparisons for the validation regimes and the full 
training-test split, respectively. Across all fold pairs, the null hypothesis of equal distributions is 
rejected at the 1\% level, indicating regime separation and supporting the use of multi-regime validation.

\begin{table}[H]
    \caption{\textbf{Pairwise Kolmogorov-Smirnov Test Results Across Market Regimes.}}\label{tab:ks_tests}
    \centering
\resizebox{\textwidth}{!}{%
\begin{tabular}{llcccc}
\toprule
& & \multicolumn{2}{c}{\textbf{Return}} & \multicolumn{2}{c}{\textbf{Volatility (5-day)}} \\
\cmidrule(lr){3-4} \cmidrule(lr){5-6}
\textbf{Regime A} & \textbf{Regime B} & \textbf{KS Statistic} & \textbf{$p$-value} & \textbf{KS Statistic} & \textbf{$p$-value} \\
\midrule
2022 & 2023 & 0.087 & $< 0.001$ & 0.238 & $< 0.001$ \\
2022 & 2024 & 0.101 & $< 0.001$ & 0.277 & $< 0.001$ \\
2022 & 2025 & 0.077 & $< 0.001$ & 0.178 & $< 0.001$ \\
2023 & 2024 & 0.015 & $< 0.001$ & 0.048 & $< 0.001$ \\
2023 & 2025 & 0.020 & $< 0.001$ & 0.061 & $< 0.001$ \\
2024 & 2025 & 0.033 & $< 0.001$ & 0.104 & $< 0.001$ \\
\midrule
\multicolumn{2}{l}{Training (2015--2024) vs OOS (2025)} & 0.022 & $< 0.001$ & 0.070 & $< 0.001$ \\
\bottomrule
\end{tabular}%
}
    \vspace{0.5em}
    \raggedright
    \footnotesize\textit{Note: All tests are statistically significant at $p < 0.001$.}
\end{table}

\section{Equity Curves}\label{app:res}

\begin{figure}[H]
    \centering
    \includegraphics[width=0.9\textwidth]{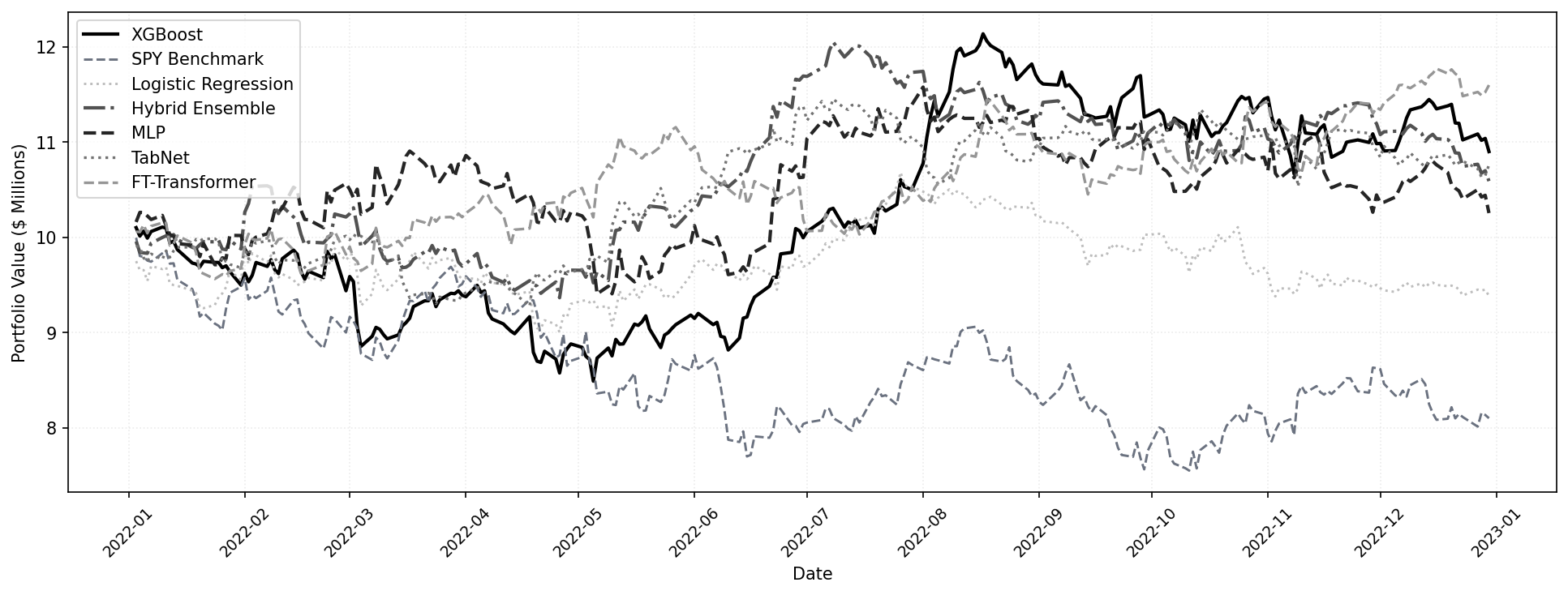}
    \caption{Equity curves for all models vs S\&P 500 benchmark, 2022 validation period.}
    \label{fig:equity_2022}
\end{figure}

\begin{figure}[H]
    \centering
    \includegraphics[width=0.9\textwidth]{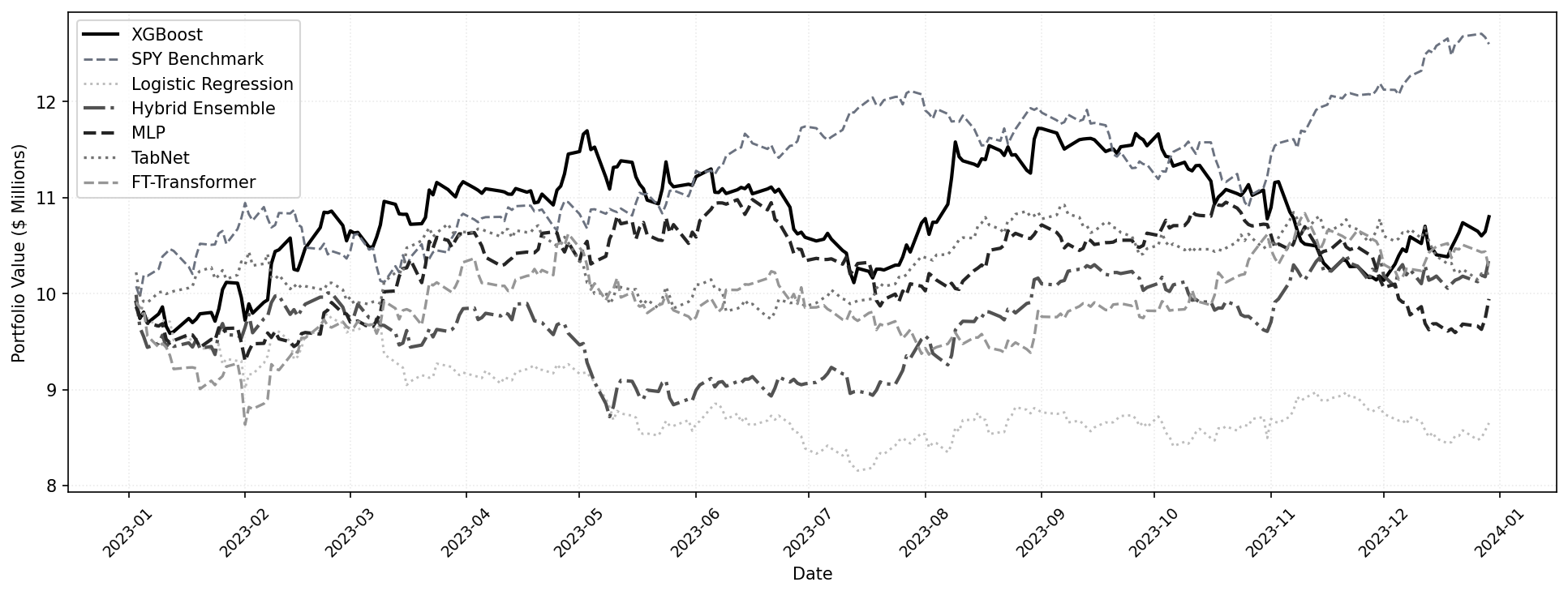}
    \caption{Equity curves for all models vs S\&P 500 benchmark, 2023 validation period.}
    \label{fig:equity_2023}
\end{figure}

\begin{figure}[H]
    \centering
    \includegraphics[width=0.9\textwidth]{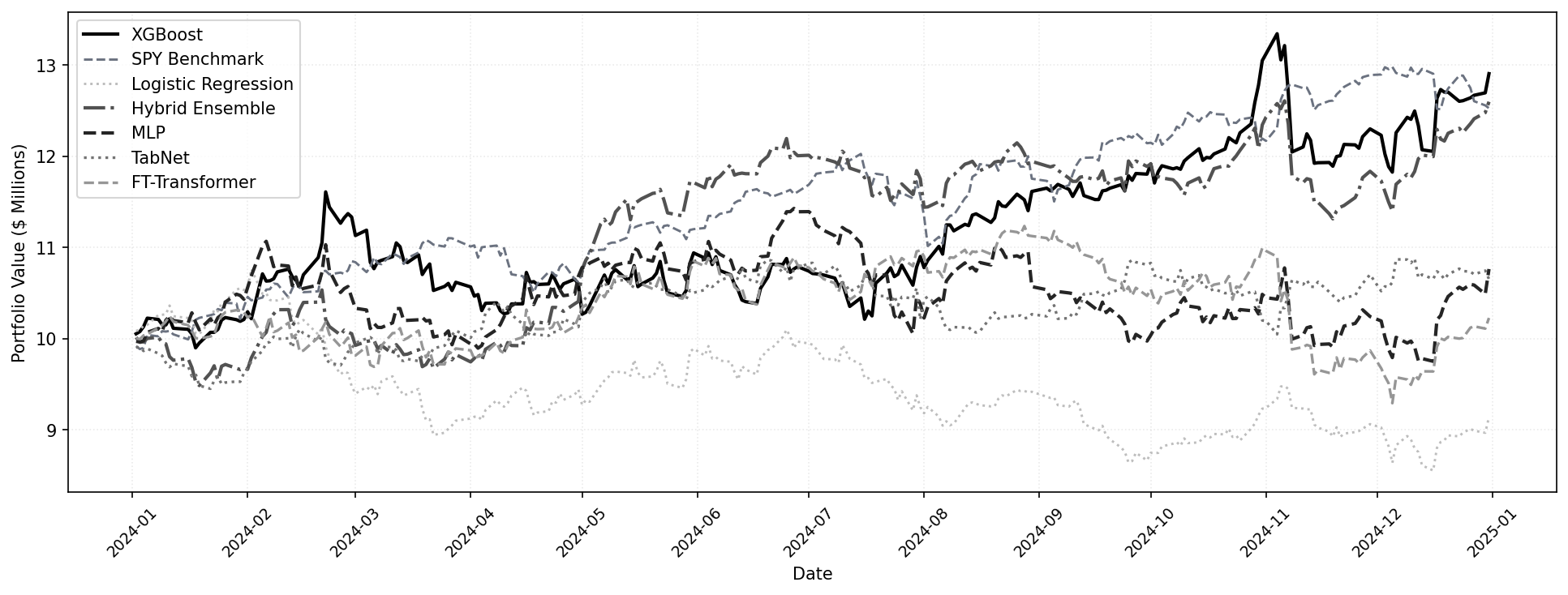}
    \caption{Equity curves for all models vs S\&P 500 benchmark, 2024 validation period.}
    \label{fig:equity_2024}
\end{figure}

Figures~\ref{fig:equity_2022}--\ref{fig:equity_2024} present equity curves for 
all models and the S\&P 500 benchmark across the three validation regimes.
All portfolios are initialised at \$10 million. 
The 2022 and 2023 periods correspond to the first and second Bayesian optimisation 
validation folds; 2024 is the final validation fold. 

\vspace{-2em}

\vspace{-1em}

\vspace{-1em}

\section{Confusion Matrices}\label{app:conf}
\begin{figure}[H]
   \input{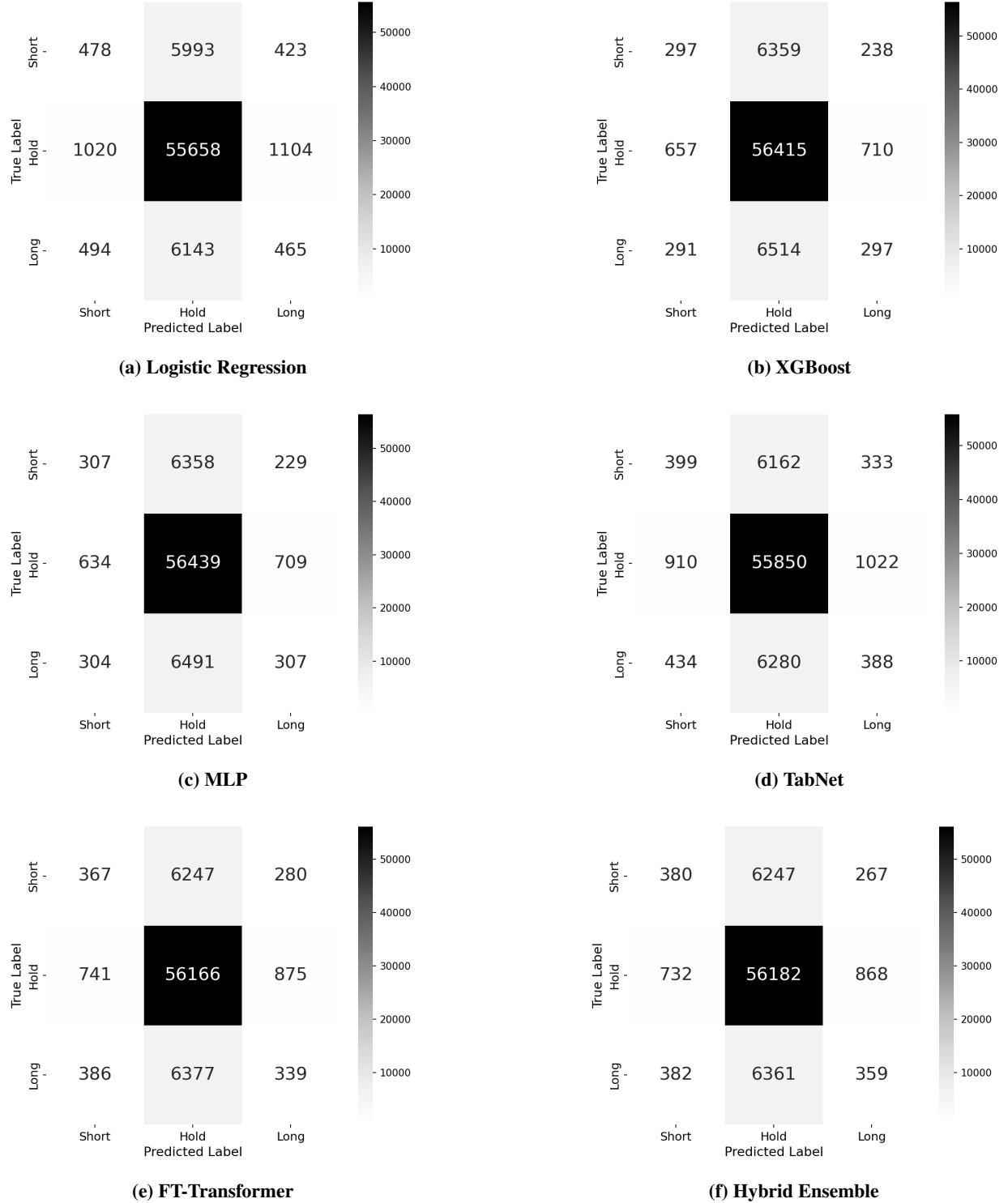}
    \caption{Confusion matrices for all models evaluated on the OOS period.}
    \label{confusion_matrices}
\end{figure}

Figure~\ref{confusion_matrices} presents confusion matrices for all 
models evaluated on the OOS 2025 period. The pattern 
across all models is majority assignment to the Hold class, consistent with the 
80\% class imbalance and top rank selection. Long and Short diagonal counts confirm 
that all models retain above-random signal in the minority classes.

\section{Statistical Test Results}
\label{app:stats}

\subsection{Shapiro--Wilk Normality Test}

Table~\ref{tab:shapiro_wilk} reports Shapiro--Wilk test results for the daily 
returns of each model and the S\&P 500 benchmark over the OOS period. 
Only LR ($W = 0.992$, $p = 0.177$) and MLP ($W = 0.989$, 
$p = 0.062$) fail to reject normality at the 5\% significance level, so 
non-parametric tests are used for all models.

\begin{table}[H]
    \centering
    \caption{Shapiro--Wilk normality test results for daily returns.}
    \label{tab:shapiro_wilk}
    \begin{tabular}{lccc}
\toprule
\textbf{Model} & \textbf{W Statistic} & \textbf{p-value} & \textbf{Normal?} \\
\midrule
SPY                 & 0.7831 & $<$0.001 & No  \\
Logistic Regression & 0.9918 & 0.177    & Yes \\
XGBoost             & 0.8739 & $<$0.001 & No  \\
MLP                 & 0.9893 & 0.062    & Yes \\
TabNet              & 0.9586 & $<$0.001 & No  \\
FT-Transformer      & 0.9404 & $<$0.001 & No  \\
Hybrid              & 0.9470 & $<$0.001 & No  \\
\bottomrule
\end{tabular}
\end{table}

\subsection{Pairwise CAPM Regressions Between Models}

Table~\ref{tab:capm_pairwise} presents pairwise CAPM regressions across all 
model pairs, where relative alpha captures return generation beyond what a 
given comparison model explains. The Hybrid produces statistically significant 
relative alphas against MLP ($p = 0.015$), TabNet ($p = 0.009$), and 
FT-Transformer ($p = 0.005$), indicating it generates excess returns that 
cannot be attributed to shared signal structure.

\begin{table}[H]
    \centering
    \small
    \caption{Pairwise CAPM regression results between models. Ann.\ $\alpha$ 
             is annualised relative alpha.}
    \label{tab:capm_pairwise}
    \renewcommand{\arraystretch}{1.3}
\begin{tabular}{lcccc}
\toprule
\textbf{Pair} & \textbf{Ann. $\alpha$} & \textbf{$p(\alpha)$} & \textbf{$\beta$} & \textbf{$R^2$} \\
\midrule
LogReg vs XGBoost         & $-$0.148 & 0.226 & 0.314 & 0.218 \\
LogReg vs Hybrid          & $-$0.230 & 0.059 & 0.416 & 0.247 \\
LogReg vs MLP             & $-$0.082 & 0.542 & 0.177 & 0.059 \\
LogReg vs TabNet          & $-$0.067 & 0.621 & 0.169 & 0.034 \\
LogReg vs FT-Transformer  & $-$0.061 & 0.645 & 0.236 & 0.088 \\
\midrule
XGBoost vs Hybrid         & $-$0.022 & 0.894 & 0.776 & 0.388 \\
XGBoost vs MLP            & 0.238    & 0.209 & 0.420 & 0.149 \\
XGBoost vs TabNet         & 0.291    & 0.151 & 0.226 & 0.028 \\
XGBoost vs FT-Transformer & 0.285    & 0.102 & 0.625 & 0.279 \\
\midrule
\textbf{Hybrid vs MLP}    & \textbf{0.373} & \textbf{0.015} & 0.331 & 0.144 \\
\textbf{Hybrid vs TabNet} & \textbf{0.377} & \textbf{0.009} & 0.550 & 0.253 \\
\textbf{Hybrid vs FT-Transformer} & \textbf{0.413} & \textbf{0.005} & 0.438 & 0.213 \\
\midrule
MLP vs TabNet             & 0.164    & 0.382 & 0.154 & 0.015 \\
MLP vs FT-Transformer     & 0.157    & 0.350 & 0.497 & 0.209 \\
\midrule
TabNet vs FT-Transformer  & 0.094    & 0.526 & 0.160 & 0.034 \\
\bottomrule
\end{tabular}
\renewcommand{\arraystretch}{1}
\end{table}

\newpage

\section{Robustness and Temporal Stability}
\label{app:robustness_figures}

\begin{figure}[H]
    \centering
    \includegraphics[width=\textwidth]{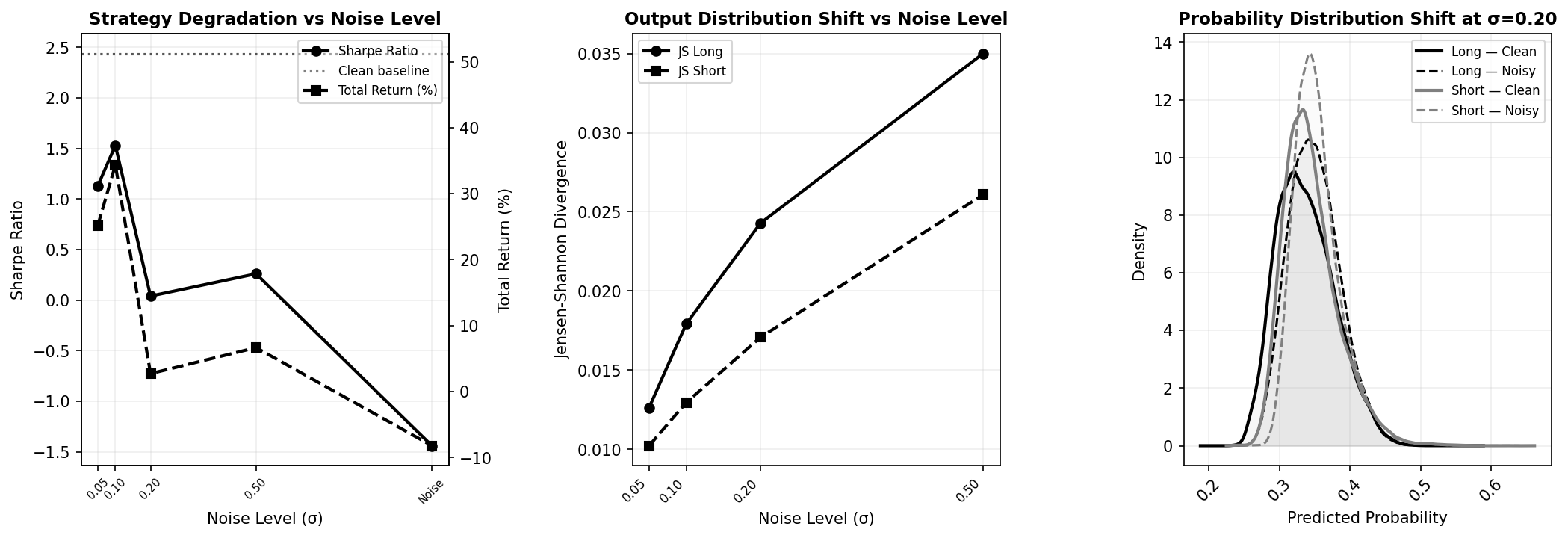}
    \caption{Hybrid model Gaussian noise robustness analysis. Left: strategy 
    degradation vs noise level. Centre: JS divergence vs noise level. 
    Right: probability distribution shift at $\sigma = 0.20$.}
    \label{fig:gaussian_robustness}
\end{figure}

\begin{figure}[H]
    \centering
    \includegraphics[width=\textwidth]{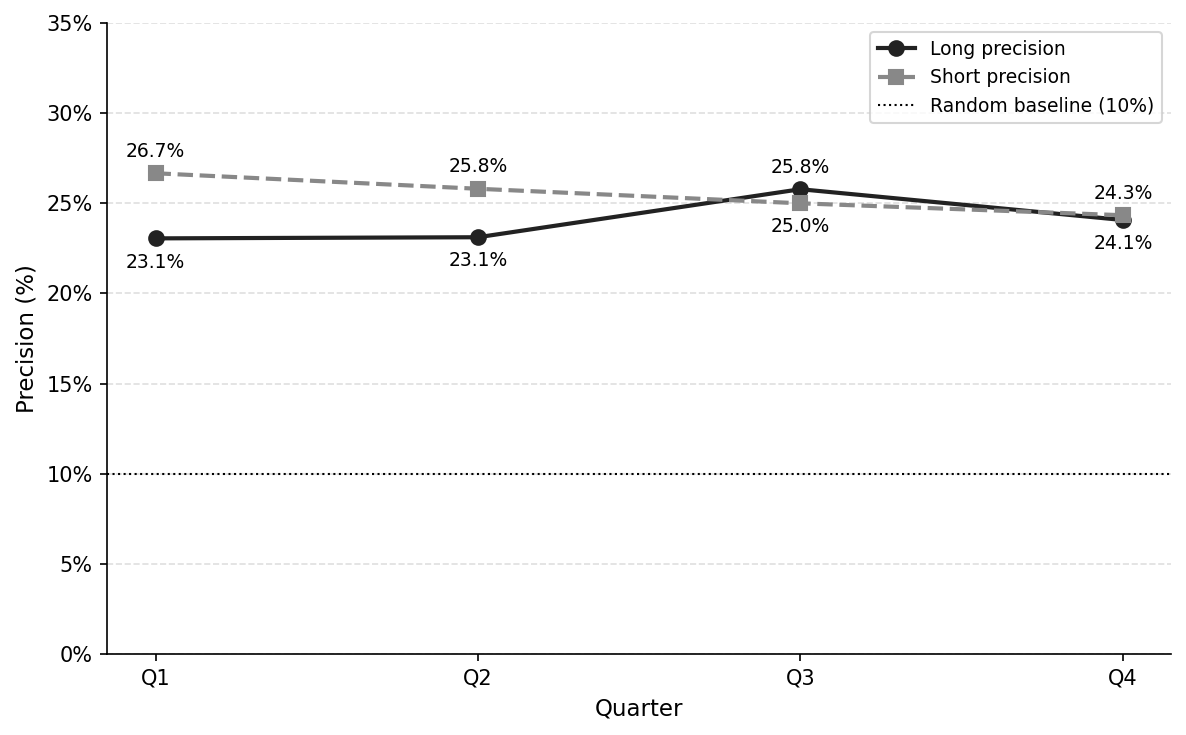}
    \caption{Hybrid model quarterly signal precision across the 2025 
    OOS period. The dotted line indicates the 10\% random baseline.}
    \label{fig:quarterly_precision}
\end{figure}

Figure~\ref{fig:gaussian_robustness} presents the full Gaussian noise robustness 
analysis for the Hybrid model, showing strategy degradation, output distribution 
shift, and the probability distribution shift at $\sigma = 0.20$. 
Figure~\ref{fig:quarterly_precision} plots Long and Short signal precision 
across the four OOS quarters against the 10\% random baseline.

\newpage

\section{Feature Importances}\label{app:shap}

This appendix presents the full per-architecture SHAP analysis from Section~\ref{sec:shap_results}. 
Table~\ref{tab:appendix-category-comparison}
compares the four feature category contributions; Technical, Fundamental,
Macroeconomic, Alternative across all five base models for both long and short
signal directions. Tables~\ref{tab:appendix-logreg}--\ref{tab:appendix-ft} report
each remaining model's individual Top 10 features by mean absolute SHAP
value, alongside its category breakdown.

\begin{table}[H]
    \centering
    \footnotesize
    \caption{Feature Category SHAP Attribution Across All Base Models}
    \label{tab:appendix-category-comparison}
    \renewcommand{\arraystretch}{1.15}
\begin{tabular*}{\textwidth}{@{\extracolsep{\fill}} l l r r r r}
\toprule
\multicolumn{2}{c}{\textbf{Model / Direction}} 
    & \textbf{Technical} & \textbf{Fundamental} & \textbf{Macroeconomic} & \textbf{Alternative} \\
\cmidrule(lr){1-2}\cmidrule(lr){3-6}
\multirow{2}{*}{XGBoost}
    & Long  & 50.87\% & 43.73\% & 3.76\%  & 1.63\% \\
    & Short & 59.45\% & 33.30\% & 4.36\%  & 2.89\% \\
\addlinespace[0.15cm]
\multirow{2}{*}{Logistic Regression}
    & Long  & 55.93\% & 14.44\% & 15.56\% & 14.08\% \\
    & Short & 63.09\% & 11.89\% & 12.82\% & 12.20\% \\
\addlinespace[0.15cm]
\multirow{2}{*}{MLP}
    & Long  & 40.94\% & 32.98\% & 13.52\% & 12.56\% \\
    & Short & 47.51\% & 27.17\% & 15.16\% & 10.16\% \\
\addlinespace[0.15cm]
\multirow{2}{*}{TabNet}
    & Long  & 58.97\% & 18.41\% & 15.43\% & 7.19\% \\
    & Short & 57.79\% & 21.60\% & 12.12\% & 8.48\% \\
\addlinespace[0.15cm]
\multirow{2}{*}{FT-Transformer}
    & Long  & 52.58\% & 30.30\% & 8.90\%  & 8.22\% \\
    & Short & 55.94\% & 27.01\% & 9.43\%  & 7.62\% \\
\bottomrule
\end{tabular*}
\begin{tablenotes}
\footnotesize
\item \textbf{Note}: Each cell represents the percentage share of total mean 
absolute SHAP attribution assigned to that feature category across the full 
50-feature set, for each model and signal direction.
\end{tablenotes}
\end{table}

\begin{table}[H]
    \centering
    \caption{Top 10 Features by Mean Absolute SHAP Value --- Logistic Regression}
    \label{tab:appendix-logreg}
    \footnotesize
\renewcommand{\arraystretch}{1.1}
\begin{minipage}[t]{0.48\textwidth}
\setlength{\tabcolsep}{0pt}
\begin{tabular*}{\textwidth}{@{\extracolsep{\fill}} l r}
\toprule
\multicolumn{2}{c}{\textbf{Long Signal}} \\
\cmidrule{1-2}
\textbf{Feature} & \textbf{Mean SHAP} \\
\midrule
\addlinespace[0.1cm]
Distance from 52-Week High  & 0.0208 \\
Distance from 50-Day MA     & 0.0091 \\
Analyst Rating Consensus    & 0.0065 \\
Relative Volume             & 0.0064 \\
5-Day Volatility            & 0.0064 \\
Bollinger Band Width        & 0.0044 \\
Core Inflation              & 0.0044 \\
Unemployment Trend          & 0.0041 \\
Yield Curve Slope           & 0.0037 \\
Investment Trend            & 0.0029 \\
\addlinespace[0.1cm]
\bottomrule
\end{tabular*}
\end{minipage}
\hfill
\begin{minipage}[t]{0.48\textwidth}
\setlength{\tabcolsep}{0pt}
\begin{tabular*}{\textwidth}{@{\extracolsep{\fill}} l r}
\toprule
\multicolumn{2}{c}{\textbf{Short Signal}} \\
\cmidrule{1-2}
\textbf{Feature} & \textbf{Mean SHAP} \\
\midrule
\addlinespace[0.1cm]
Distance from 52-Week High  & 0.0190 \\
Distance from 50-Day MA     & 0.0086 \\
5-Day Volatility            & 0.0073 \\
Relative Volume             & 0.0066 \\
Relative RSI                & 0.0043 \\
Core Inflation              & 0.0040 \\
Bollinger Band Width        & 0.0038 \\
Return on Assets            & 0.0037 \\
Unemployment Trend          & 0.0036 \\
Return Rank (Percentile)    & 0.0035 \\
\addlinespace[0.1cm]
\bottomrule
\end{tabular*}
\end{minipage}

\vspace{0.8em}

\begin{tabular*}{\textwidth}{@{\extracolsep{\fill}} l r r}
\toprule
\multicolumn{3}{c}{\textbf{Total Mean SHAP Contribution by Feature Category (All Features)}} \\
\cmidrule{1-3}
\textbf{Category} & \textbf{Buy Signal} & \textbf{Sell Signal} \\
\midrule
Technical       & 55.93\% & 63.09\% \\
Fundamental     & 14.44\% & 11.89\% \\
Macroeconomic   & 15.56\% & 12.82\% \\
Alternative     & 14.08\% & 12.20\% \\
\bottomrule
\end{tabular*}

\begin{tablenotes}
\footnotesize
\item \textbf{Note}: Top 10 features ranked by mean absolute SHAP value for buy 
and sell signal directions, shown for Logistic Regression. 
Features may differ between directions. Category totals represent each 
category's share of total mean absolute SHAP attribution across the full 
50-feature set for this model. Per-architecture breakdowns for all five base 
models are provided in Appendix~\ref{app:shap}.
\end{tablenotes}
\end{table}

\begin{table}[H]
    \centering
    \caption{Top 10 Features by Mean Absolute SHAP Value --- XGBoost}
    \label{tab:appendix-xgb}
    \footnotesize
\renewcommand{\arraystretch}{1.1}
\begin{minipage}[t]{0.48\textwidth}
\setlength{\tabcolsep}{0pt}
\begin{tabular*}{\textwidth}{@{\extracolsep{\fill}} l r}
\toprule
\multicolumn{2}{c}{\textbf{Long Signal}} \\
\cmidrule{1-2}
\textbf{Feature} & \textbf{Mean SHAP} \\
\midrule
\addlinespace[0.1cm]
Market Capitalisation & 0.0216 \\
Trading Volume & 0.0116 \\
Daily Dollar Volume & 0.0086 \\
5-Day Volatility & 0.0083 \\
Distance from 52-Week High & 0.0068 \\
Diluted EPS & 0.0047 \\
Price-to-Cash-Flow Ratio & 0.0045 \\
Relative Volume & 0.0040 \\
Analyst Rating Consensus & 0.0040 \\
Return Rank (Percentile) & 0.0039 \\
\addlinespace[0.1cm]
\bottomrule
\end{tabular*}
\end{minipage}
\hfill
\begin{minipage}[t]{0.48\textwidth}
\setlength{\tabcolsep}{0pt}
\begin{tabular*}{\textwidth}{@{\extracolsep{\fill}} l r}
\toprule
\multicolumn{2}{c}{\textbf{Short Signal}} \\
\cmidrule{1-2}
\textbf{Feature} & \textbf{Mean SHAP} \\
\midrule
\addlinespace[0.1cm]
Trading Volume & 0.0092 \\
Daily Dollar Volume & 0.0078 \\
5-Day Volatility & 0.0077 \\
Market Capitalisation & 0.0074 \\
Distance from 52-Week High & 0.0070 \\
Price-to-Cash-Flow Ratio & 0.0065 \\
Return Rank (Percentile) & 0.0055 \\
Shares Outstanding & 0.0045 \\
5-Day Relative Return & 0.0029 \\
Relative Volume & 0.0024 \\
\addlinespace[0.1cm]
\bottomrule
\end{tabular*}
\end{minipage}

\vspace{0.8em}

\begin{tabular*}{\textwidth}{@{\extracolsep{\fill}} l r r}
\toprule
\multicolumn{3}{c}{\textbf{Total Mean SHAP Contribution by Feature Category (All Features)}} \\
\cmidrule{1-3}
\textbf{Category} & \textbf{Buy Signal} & \textbf{Sell Signal} \\
\midrule
Technical       & 50.87\% & 59.45\% \\
Fundamental     & 43.73\% & 33.30\% \\
Macroeconomic   & 3.76\%  & 4.36\%  \\
Alternative     & 1.63\%  & 2.89\%  \\
\bottomrule
\end{tabular*}

\begin{tablenotes}
\footnotesize
\item \textbf{Note}: Top 10 features ranked by mean absolute SHAP value for buy 
and sell signal directions, shown for the best-performing individual model 
(XGBoost). Features may differ between directions. Category 
totals represent each category's share of total mean absolute SHAP attribution 
across the full 50-feature set for this model. Per-architecture breakdowns for 
all five base models are provided in Appendix~\ref{app:shap}.
\end{tablenotes}
\end{table}

\begin{table}[H]
    \centering
    \caption{Top 10 Features by Mean Absolute SHAP Value --- MLP}
    \label{tab:appendix-mlp}
    \footnotesize
\renewcommand{\arraystretch}{1.1}
\begin{minipage}[t]{0.48\textwidth}
\setlength{\tabcolsep}{0pt}
\begin{tabular*}{\textwidth}{@{\extracolsep{\fill}} l r}
\toprule
\multicolumn{2}{c}{\textbf{Long Signal}} \\
\cmidrule{1-2}
\textbf{Feature} & \textbf{Mean SHAP} \\
\midrule
\addlinespace[0.1cm]
Trading Volume              & 0.0230 \\
Market Capitalisation       & 0.0215 \\
Shares Outstanding          & 0.0132 \\
Distance from 52-Week High  & 0.0087 \\
Daily Dollar Volume         & 0.0083 \\
Analyst Rating Consensus    & 0.0081 \\
Investment Trend            & 0.0051 \\
Price-to-Book Ratio         & 0.0046 \\
Core Inflation              & 0.0042 \\
5-Day Volatility            & 0.0039 \\
\addlinespace[0.1cm]
\bottomrule
\end{tabular*}
\end{minipage}
\hfill
\begin{minipage}[t]{0.48\textwidth}
\setlength{\tabcolsep}{0pt}
\begin{tabular*}{\textwidth}{@{\extracolsep{\fill}} l r}
\toprule
\multicolumn{2}{c}{\textbf{Short Signal}} \\
\cmidrule{1-2}
\textbf{Feature} & \textbf{Mean SHAP} \\
\midrule
\addlinespace[0.1cm]
Daily Dollar Volume         & 0.0181 \\
Shares Outstanding          & 0.0109 \\
Market Capitalisation       & 0.0109 \\
Trading Volume              & 0.0099 \\
Distance from 52-Week High  & 0.0065 \\
Distance from 50-Day MA     & 0.0043 \\
Price-to-Sales Ratio        & 0.0040 \\
5-Day Relative Return       & 0.0040 \\
20-Day Relative Return      & 0.0037 \\
Return on Assets            & 0.0036 \\
\addlinespace[0.1cm]
\bottomrule
\end{tabular*}
\end{minipage}

\vspace{0.8em}

\begin{tabular*}{\textwidth}{@{\extracolsep{\fill}} l r r}
\toprule
\multicolumn{3}{c}{\textbf{Total Mean SHAP Contribution by Feature Category (All Features)}} \\
\cmidrule{1-3}
\textbf{Category} & \textbf{Buy Signal} & \textbf{Sell Signal} \\
\midrule
Technical       & 40.94\% & 47.51\% \\
Fundamental     & 32.98\% & 27.17\% \\
Macroeconomic   & 13.52\% & 15.16\% \\
Alternative     & 12.56\% & 10.16\% \\
\bottomrule
\end{tabular*}

\begin{tablenotes}
\footnotesize
\item \textbf{Note}: Top 10 features ranked by mean absolute SHAP value for buy 
and sell signal directions, shown for the MLP. Features may differ 
between directions. Category totals represent each category's share of total 
mean absolute SHAP attribution across the full 50-feature set for this model. 
Per-architecture breakdowns for all five base models are provided in 
Appendix~\ref{app:shap}.
\end{tablenotes}
\end{table}

\begin{table}[H]
    \centering
    \caption{Top 10 Features by Mean Absolute SHAP Value --- TabNet}
    \label{tab:appendix-tabnet}
    \footnotesize
\renewcommand{\arraystretch}{1.1}
\begin{minipage}[t]{0.48\textwidth}
\setlength{\tabcolsep}{0pt}
\begin{tabular*}{\textwidth}{@{\extracolsep{\fill}} l r}
\toprule
\multicolumn{2}{c}{\textbf{Long Signal}} \\
\cmidrule{1-2}
\textbf{Feature} & \textbf{Mean SHAP} \\
\midrule
\addlinespace[0.1cm]
Distance from 52-Week High  & 0.0131 \\
5-Day Volatility            & 0.0089 \\
Relative Volume             & 0.0073 \\
VIX Regime (High)           & 0.0057 \\
Bollinger Band Width        & 0.0038 \\
Diluted EPS                 & 0.0034 \\
Price-to-Sales Ratio        & 0.0029 \\
Return on Assets            & 0.0027 \\
Shares Outstanding          & 0.0016 \\
Unemployment Trend          & 0.0013 \\
\addlinespace[0.1cm]
\bottomrule
\end{tabular*}
\end{minipage}
\hfill
\begin{minipage}[t]{0.48\textwidth}
\setlength{\tabcolsep}{0pt}
\begin{tabular*}{\textwidth}{@{\extracolsep{\fill}} l r}
\toprule
\multicolumn{2}{c}{\textbf{Short Signal}} \\
\cmidrule{1-2}
\textbf{Feature} & \textbf{Mean SHAP} \\
\midrule
\addlinespace[0.1cm]
Distance from 52-Week High  & 0.0090 \\
5-Day Volatility            & 0.0069 \\
Relative Volume             & 0.0051 \\
Return on Assets            & 0.0033 \\
Bollinger Band Width        & 0.0025 \\
Price-to-Sales Ratio        & 0.0024 \\
VIX Regime (High)           & 0.0024 \\
Diluted EPS                 & 0.0022 \\
Shares Outstanding          & 0.0020 \\
Inflation Trend             & 0.0011 \\
\addlinespace[0.1cm]
\bottomrule
\end{tabular*}
\end{minipage}

\vspace{0.8em}

\begin{tabular*}{\textwidth}{@{\extracolsep{\fill}} l r r}
\toprule
\multicolumn{3}{c}{\textbf{Total Mean SHAP Contribution by Feature Category (All Features)}} \\
\cmidrule{1-3}
\textbf{Category} & \textbf{Buy Signal} & \textbf{Sell Signal} \\
\midrule
Technical       & 58.97\% & 57.79\% \\
Fundamental     & 18.41\% & 21.60\% \\
Macroeconomic   & 15.43\% & 12.12\% \\
Alternative     & 7.19\%  & 8.48\%  \\
\bottomrule
\end{tabular*}

\begin{tablenotes}
\footnotesize
\item \textbf{Note}: Top 10 features ranked by mean absolute SHAP value for buy 
and sell signal directions, shown for TabNet. Features may differ 
between directions. Category totals represent each category's share of total 
mean absolute SHAP attribution across the full 50-feature set for this model. 
Per-architecture breakdowns for all five base models are provided in 
Appendix~\ref{app:shap}.
\end{tablenotes}
\end{table}

\begin{table}[H]
    \centering
    \caption{Top 10 Features by Mean Absolute SHAP Value --- FT-Transformer}
    \label{tab:appendix-ft}
    \footnotesize
\renewcommand{\arraystretch}{1.1}
\begin{minipage}[t]{0.48\textwidth}
\setlength{\tabcolsep}{0pt}
\begin{tabular*}{\textwidth}{@{\extracolsep{\fill}} l r}
\toprule
\multicolumn{2}{c}{\textbf{Long Signal}} \\
\cmidrule{1-2}
\textbf{Feature} & \textbf{Mean SHAP} \\
\midrule
\addlinespace[0.1cm]
Market Capitalisation       & 0.0197 \\
Daily Dollar Volume         & 0.0161 \\
Distance from 52-Week High  & 0.0090 \\
Trading Volume              & 0.0079 \\
Relative Volume             & 0.0046 \\
5-Day Volatility            & 0.0042 \\
Analyst Rating Consensus    & 0.0035 \\
Core Inflation              & 0.0033 \\
Shares Outstanding          & 0.0031 \\
Return Rank (Percentile)    & 0.0031 \\
\addlinespace[0.1cm]
\bottomrule
\end{tabular*}
\end{minipage}
\hfill
\begin{minipage}[t]{0.48\textwidth}
\setlength{\tabcolsep}{0pt}
\begin{tabular*}{\textwidth}{@{\extracolsep{\fill}} l r}
\toprule
\multicolumn{2}{c}{\textbf{Short Signal}} \\
\cmidrule{1-2}
\textbf{Feature} & \textbf{Mean SHAP} \\
\midrule
\addlinespace[0.1cm]
Daily Dollar Volume         & 0.0153 \\
Market Capitalisation       & 0.0113 \\
Distance from 52-Week High  & 0.0089 \\
Return on Assets            & 0.0052 \\
Trading Volume              & 0.0044 \\
Distance from 50-Day MA     & 0.0041 \\
5-Day Relative Return       & 0.0033 \\
Price-to-Sales Ratio        & 0.0031 \\
Return Rank (Percentile)    & 0.0031 \\
Shares Outstanding          & 0.0028 \\
\addlinespace[0.1cm]
\bottomrule
\end{tabular*}
\end{minipage}

\vspace{0.8em}

\begin{tabular*}{\textwidth}{@{\extracolsep{\fill}} l r r}
\toprule
\multicolumn{3}{c}{\textbf{Total Mean SHAP Contribution by Feature Category (All Features)}} \\
\cmidrule{1-3}
\textbf{Category} & \textbf{Buy Signal} & \textbf{Sell Signal} \\
\midrule
Technical       & 52.58\% & 55.94\% \\
Fundamental     & 30.30\% & 27.01\% \\
Macroeconomic   & 8.90\%  & 9.43\%  \\
Alternative     & 8.22\%  & 7.62\%  \\
\bottomrule
\end{tabular*}

\begin{tablenotes}
\footnotesize
\item \textbf{Note}: Top 10 features ranked by mean absolute SHAP value for buy 
and sell signal directions, shown for the FT-Transformer.
Features may differ between directions. Category totals represent each 
category's share of total mean absolute SHAP attribution across the full 
50-feature set for this model. Per-architecture breakdowns for all five base 
models are provided in Appendix~\ref{app:shap}.
\end{tablenotes}
\end{table}

\newpage

\section{Tuning, Training, and Inference Time}\label{app:runtime}

Table~\ref{tab:runtime} reports tuning, final training, and inference 
time for all models, relevant to the retraining cadence discussed in 
Section~\ref{sec:deployment}. Tuning time shows the full 30-trial 
Optuna search. Final training time represents retraining on the complete 
2015--2024 period. Inference time reflects prediction on the full 2025 
OOS set. The Hybrid ensemble is excluded from tuning and 
training figures as rank aggregation combines the pre-trained outputs 
of its constituent models without the need for a separate fitting step.

\begin{table}[H]
\centering
\caption{Tuning, Training, and Inference Time by Model}
\label{tab:runtime}
\begin{tabular}{lrrr}
\toprule
Model & Tuning (30 trials) & Final Training & OOS Inference \\
\midrule
LR & 32.0 min & 22.9 s & 0.04 s \\
XGBoost & 10.8 min & 8.4 s & 0.09 s \\
MLP & 45.0 min & 42.3 s & 2.18 s \\
TabNet & 19.6 hrs & 15.5 min & 14.0 s \\
FT-Transformer & 2.9 hrs & 3.2 min & 8.8 s \\
\bottomrule
\end{tabular}
\end{table}

\section{Interactive Application}\label{app:application}

\begin{figure}[H]
    \centering
    \includegraphics[width=\textwidth]{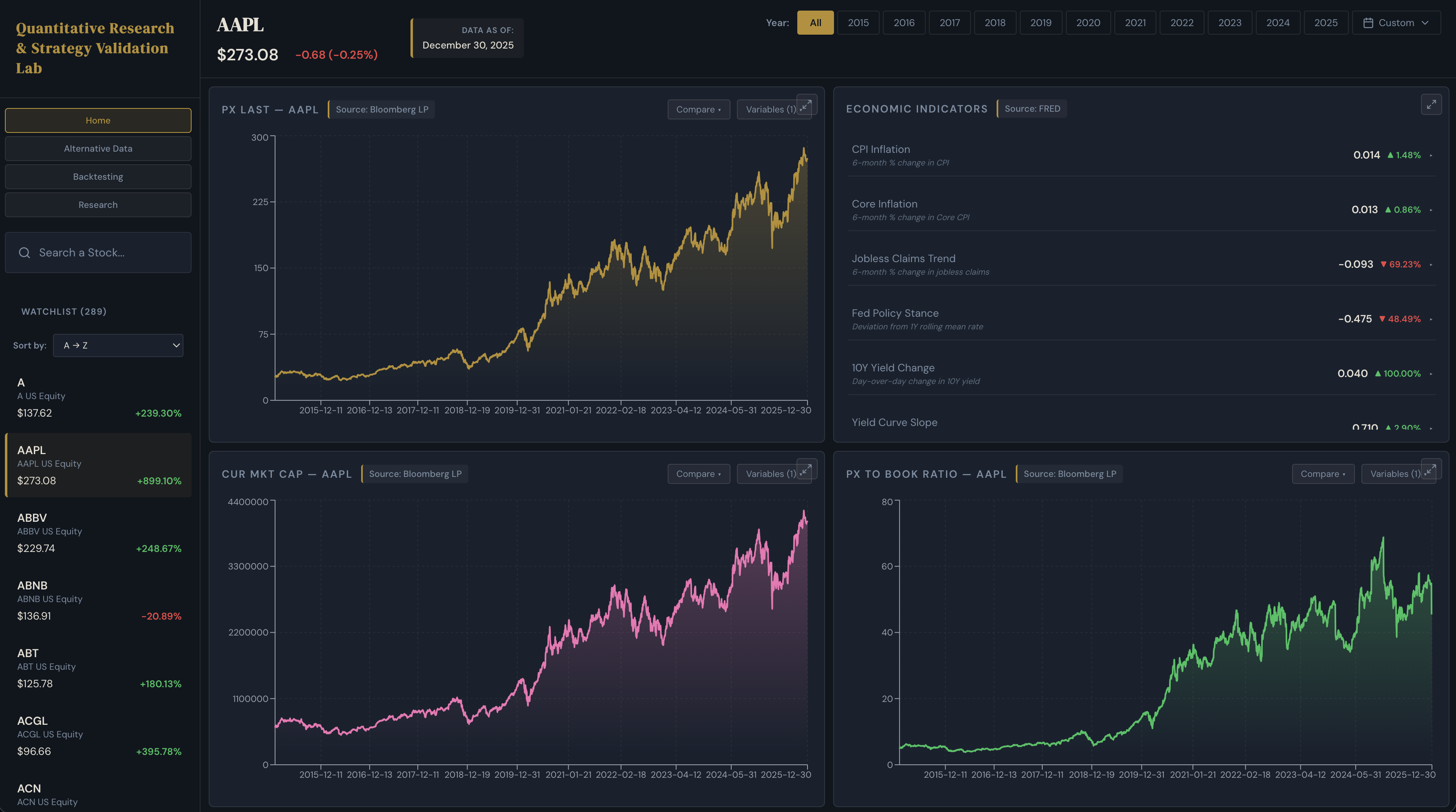}
    \caption{Stock Data View}
    \footnotesize
    Price history, market capitalisation, 
    valuation ratios, and live macroeconomic indicators for a selected 
    constituent.
    \label{fig:app_home}
\end{figure}

\begin{figure}[H]
    \centering
    \includegraphics[width=\textwidth]{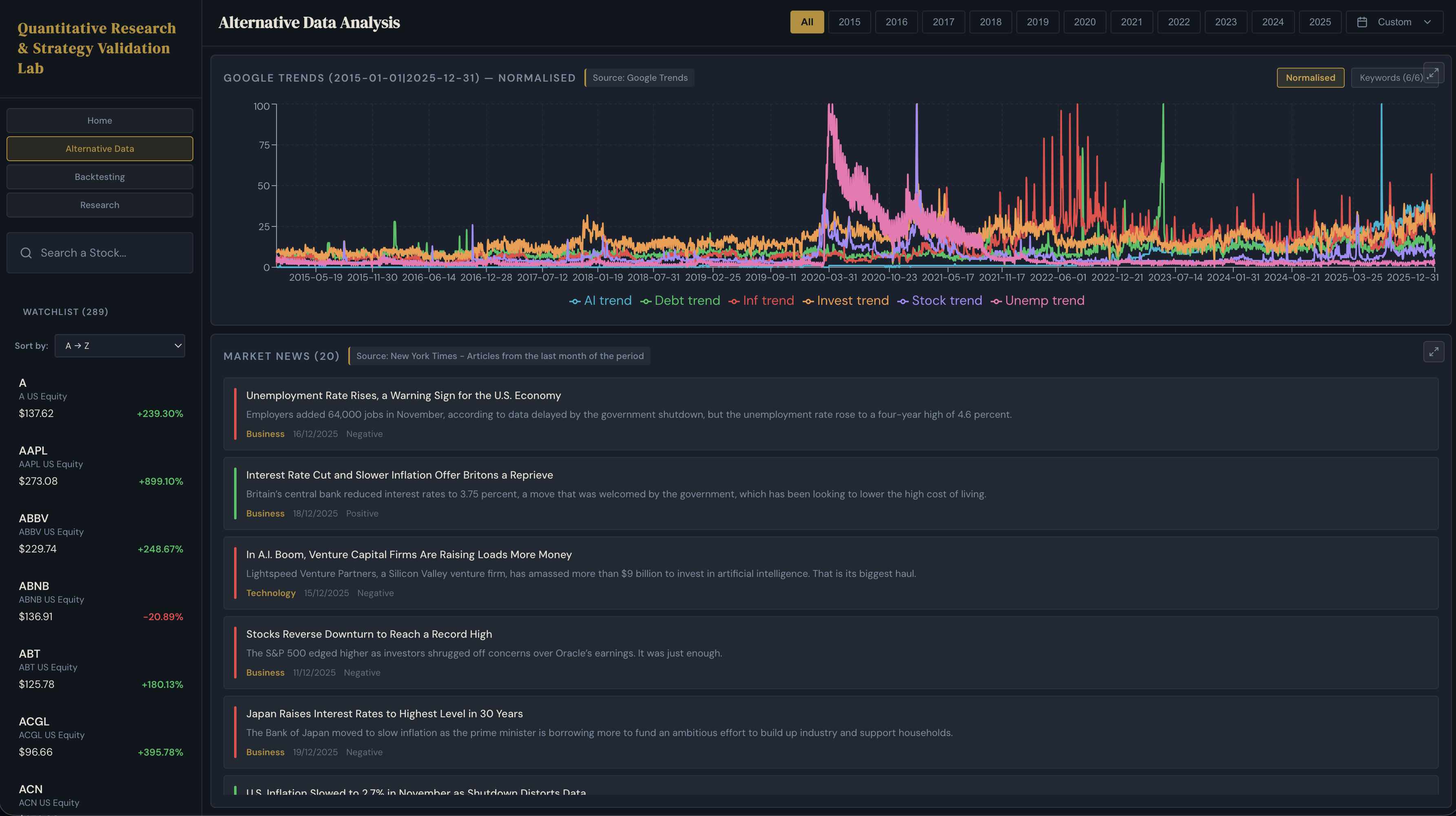}
    \caption{Alternative Data Explorer} 
    \footnotesize
    Normalised Google Trends indices 
    across the full 2015--2025 sample period, alongside firm-level news headlines with sentiment scores,
    sourced from the New York Times.
    \label{fig:app_altdata}
\end{figure}

\begin{figure}[H]
    \centering
    \includegraphics[width=\textwidth]{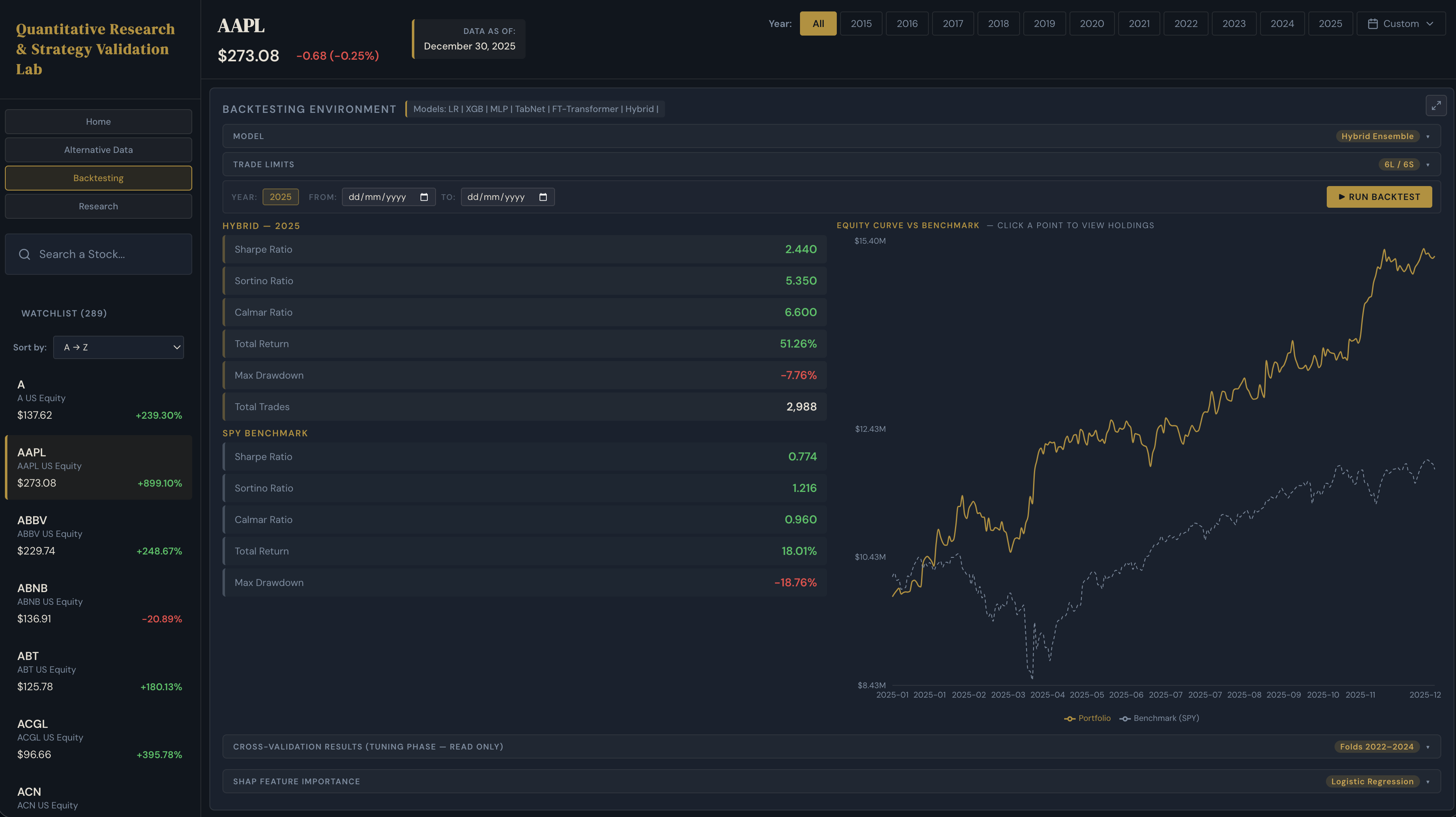}
    \caption{Interactive Backtesting Environment}
    \footnotesize
    Simulation of the long-short strategy across all six models with adjustable 
    position limits and custom date ranges, shown here for the Hybrid 
    Ensemble on the OOS period.
    \label{fig:app_backtest}
\end{figure}

\begin{figure}[H]
    \centering
    \includegraphics[width=\textwidth]{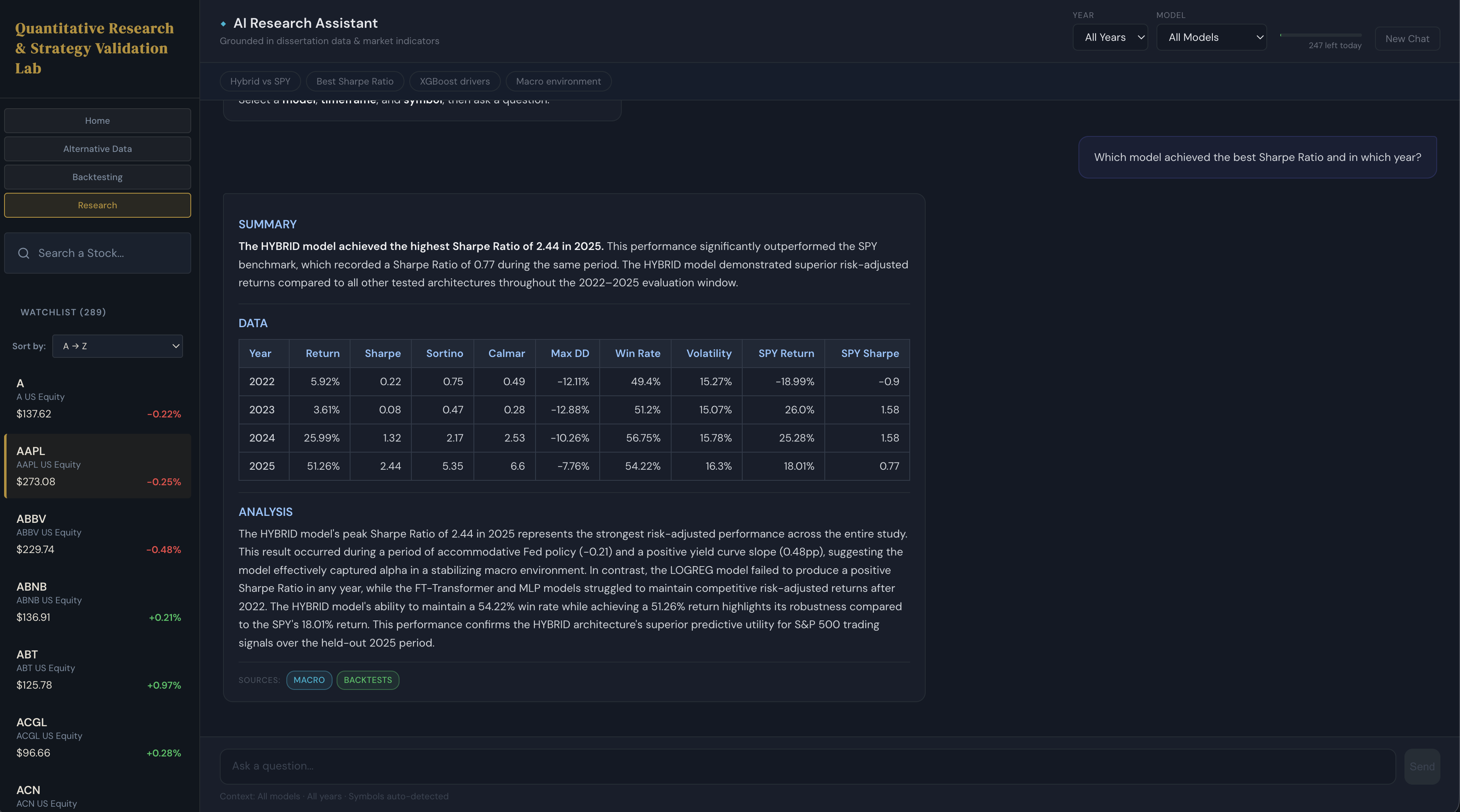}
    \caption{AI Research Assistant}
    \footnotesize
    Natural language query interface 
    grounded in the study's backtest results and macroeconomic data, 
    shown responding to a query about best Sharpe Ratio across models 
    and years.
    \label{fig:app_ai}
\end{figure}

\begin{figure}[H]
    \centering
    \resizebox{0.95\textwidth}{!}{
        \begin{tikzpicture}[
    node distance=1.2cm and 0cm,
    every node/.style={font=\fontsize{9}{11}\selectfont\sffamily},
    box/.style={rectangle, draw=black!70, thin, rounded corners=4pt,
                align=center, minimum height=0.6cm},
    smallbox/.style={rectangle, draw=black!60, thin, rounded corners=3pt,
                    align=center, fill=black!6, minimum height=0.8cm, minimum width=2.5cm},
    container/.style={rectangle, draw=black!70, thin, rounded corners=5pt,
                      fill=black!10, inner sep=4pt},
    arrow/.style={-{Latex[length=2.5mm]}, thin, black!70},
    dasharrow/.style={-{Latex[length=2.5mm]}, thin, black!50, dashed},
]

\node[box, fill=black!5, minimum width=9.5cm] (question) {
    \textbf{User Question - React Frontend}
};

\node[box, fill=black!8, minimum width=9.5cm, below=0.6cm of question] (classify) {
    \textbf{Question Classification - Gemini Flash Lite}
};
\draw[arrow] (question.south) -- (classify.north);

\node[container, below=0.6cm of classify, minimum width=9.5cm] (retrieval) {
    \begin{tabular}{c}
        \textbf{Conditional Context Retrieval} \\[2pt]
        \begin{tikzpicture}
            \node[smallbox] (c1) {Backtest};
            \node[smallbox, right=0.2cm of c1] (c2) {SHAP};
            \node[smallbox, right=0.2cm of c2] (c3) {Stock};
            \node[smallbox, below=0.2cm of c1] (c4) {Macro};
            \node[smallbox, below=0.2cm of c2] (c5) {News};
            \node[smallbox, below=0.2cm of c3] (c6) {Trends};
        \end{tikzpicture}
    \end{tabular}
};
\draw[arrow] (classify.south) -- (retrieval.north);

\node[box, fill=black!14, minimum width=4.5cm, right=1.3cm of retrieval] (sources) {
    \textbf{Cached Data} \\
    \textit{GitHub Data Store}
};
\draw[dasharrow] (sources.west) -- (retrieval.east);

\node[box, fill=black!8, minimum width=9.5cm, below=0.6cm of retrieval] (assembly) {
    \textbf{Prompt Assembly} \\
    \textit{System prompt + retrieved context}
};
\draw[arrow] (retrieval.south) -- (assembly.north);

\node[box, fill=black!14, minimum width=9.5cm, below=0.6cm of assembly] (generate) {
    \textbf{Answer Generation - Gemini Flash Lite}
};
\draw[arrow] (assembly.south) -- (generate.north);

\node[box, fill=black!5, minimum width=9.5cm, below=0.6cm of generate] (clean) {
    \textbf{Response Cleaning} \\
    \textit{Strip formatting + extract summary}
};
\draw[arrow] (generate.south) -- (clean.north);

\node[box, fill=black!5, minimum width=9.5cm, below=0.6cm of clean] (display) {
    \textbf{Displayed to User - React Frontend}
};
\draw[arrow] (clean.south) -- (display.north);

\end{tikzpicture}
    }
    \caption{AI Research Assistant Architecture}\label{fig:ai_arch}
    \footnotesize
    The assistant classifies each question to determine which context 
    types are required, retrieves only that context from the data store, 
    and generates its response only using the assembled context before returning 
    an answer to the frontend.
\end{figure}
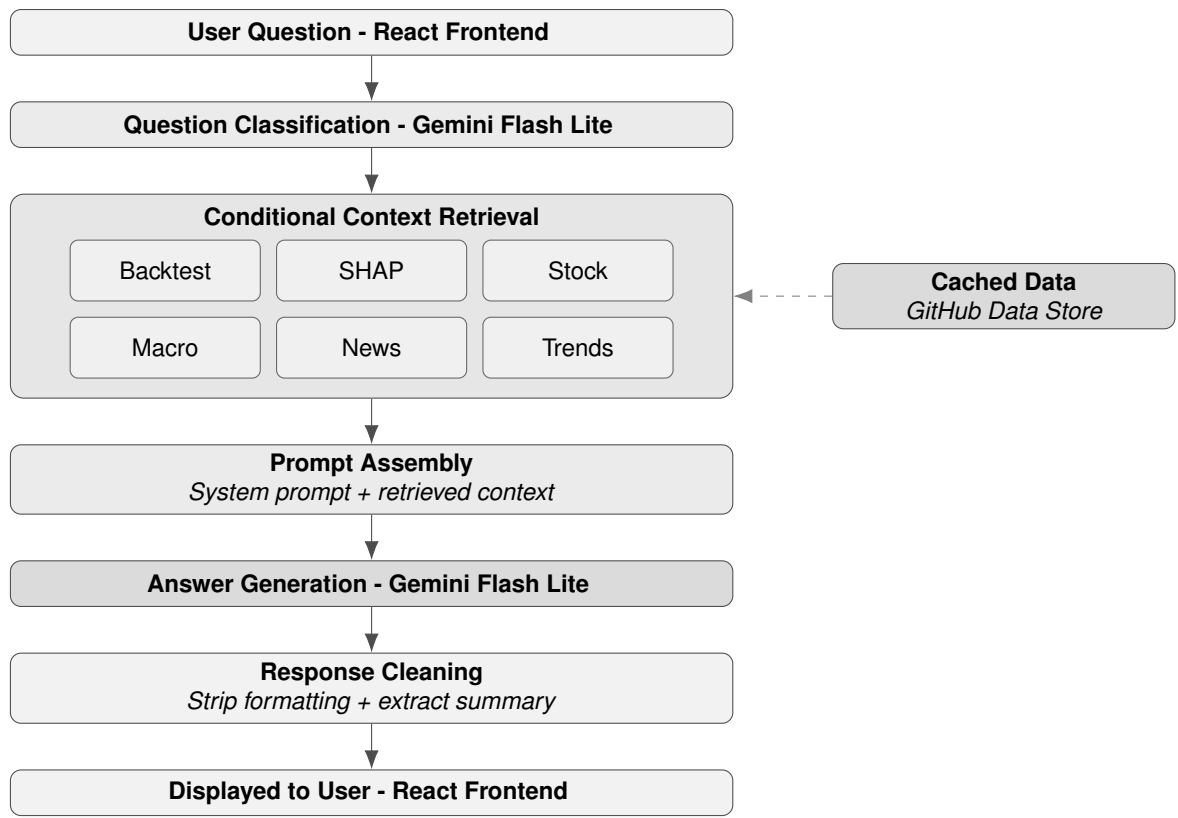

As described in Section~\ref{sec:interpretability}, a React
frontend served by a REST API was developed to operationalise the 
study's outputs, the deployed application is
accessible at \url{https://m-sc-dissertation-dev.vercel.app/}. The system exposes four 
primary views. Figure~\ref{fig:app_home} shows a stock-level data view. 
Figure~\ref{fig:app_altdata} illustrates an alternative data explorer presenting Google Trends 
indices and firm-level news sentiment across 2015--2025. 
Figure~\ref{fig:app_backtest} provides an interactive backtesting environment 
supporting real-time strategy simulation across all six models.
Figure~\ref{fig:app_ai} shows an AI research assistant 
grounded in the study's results, enabling natural language queries 
over model performance and market conditions, with its 
architecture illustrated in Figure~\ref{fig:ai_arch}.

\end{document}